\documentclass[10pt,twocolumn,letterpaper]{article}

\usepackage[pagenumbers]{cvpr}    

\usepackage{multirow}   
\usepackage{tabularx}   
\usepackage{microtype}  

\definecolor{cvprblue}{rgb}{0.21,0.49,0.74}
\usepackage[pagebackref,breaklinks,colorlinks,allcolors=cvprblue]{hyperref}

\def\confName{CVPR}
\def\confYear{2026}

\title{DiVid: Diagnosing Dimension-Specific Diversity Collapse \\ in Video Generation Models}

\author{
Huanran Hu$^{1*}$,
Zihui Ren$^{2*}$,
Dingyi Yang$^{3*}$,
Zhinan Song$^{1}$,
Guozheng Wu$^{1}$,
Tiezheng Ge$^{2}$,
Qin Jin$^{1\dagger}$ \\
$^{1}$Renmin University of China \quad
$^{2}$Alibaba Group \quad
$^{3}$Tianjin University \\
{\tt\small \{huanranhu, qjin\}@ruc.edu.cn} \\
{\tt\small \{renzihui.rzh, tiezheng.gtz\}@taobao.com} \\
}

\begin{document}
\maketitle

\begin{figure}[b]
\footnotesize
\hrule height 0.4pt width 2in
\vspace{2.6pt}
$^{*}$Equal contribution. \par
$^{\dagger}$Corresponding author.
\end{figure}

\begin{abstract}
Despite remarkable progress, video generation models often produce highly similar outputs when repeatedly sampled from the same prompt, limiting their usefulness for creative exploration. Existing diversity evaluations primarily rely on global scalar metrics, which obscure where diversity collapses in the spatiotemporal space of videos. We introduce \textbf{DiVid}, a dimension-level diagnostic framework that decomposes video generation diversity into six interpretable dimensions: Semantic, Style, Subject, Scene, Motion, and Camera. Each dimension is measured through a reproducible computer-vision pipeline and analyzed alongside quality and instruction faithfulness to examine potential trade-offs. Systematic evaluation of representative video generation models reveals that diversity is highly dimension-specific: models with strong global diversity scores still collapse on specific factors, particularly Motion and Camera. These rankings persist after filtering unfaithful generations, indicating genuine capability differences rather than off-prompt outputs. Beyond measurement, controlled prompt interventions identify two fundamental bottlenecks: default mode convergence, where models fall back to dominant patterns under open-ended prompts; and realization gaps, where models fail to faithfully realize diverse, explicitly requested alternatives, particularly for temporal factors. The larger faithfulness losses for temporal factors highlight the difficulty of controlling motion and camera variation through text alone. DiVid thus shifts the study of diversity from measuring whether it exists to diagnosing where and why it collapses, and provides actionable directions for dimension-aware training objectives and control signals. The framework will be released to facilitate future research on diverse and controllable video generation.
\end{abstract}

\section{Introduction}

\begin{figure*}[t]
    \centering
\includegraphics[width=0.95\textwidth]{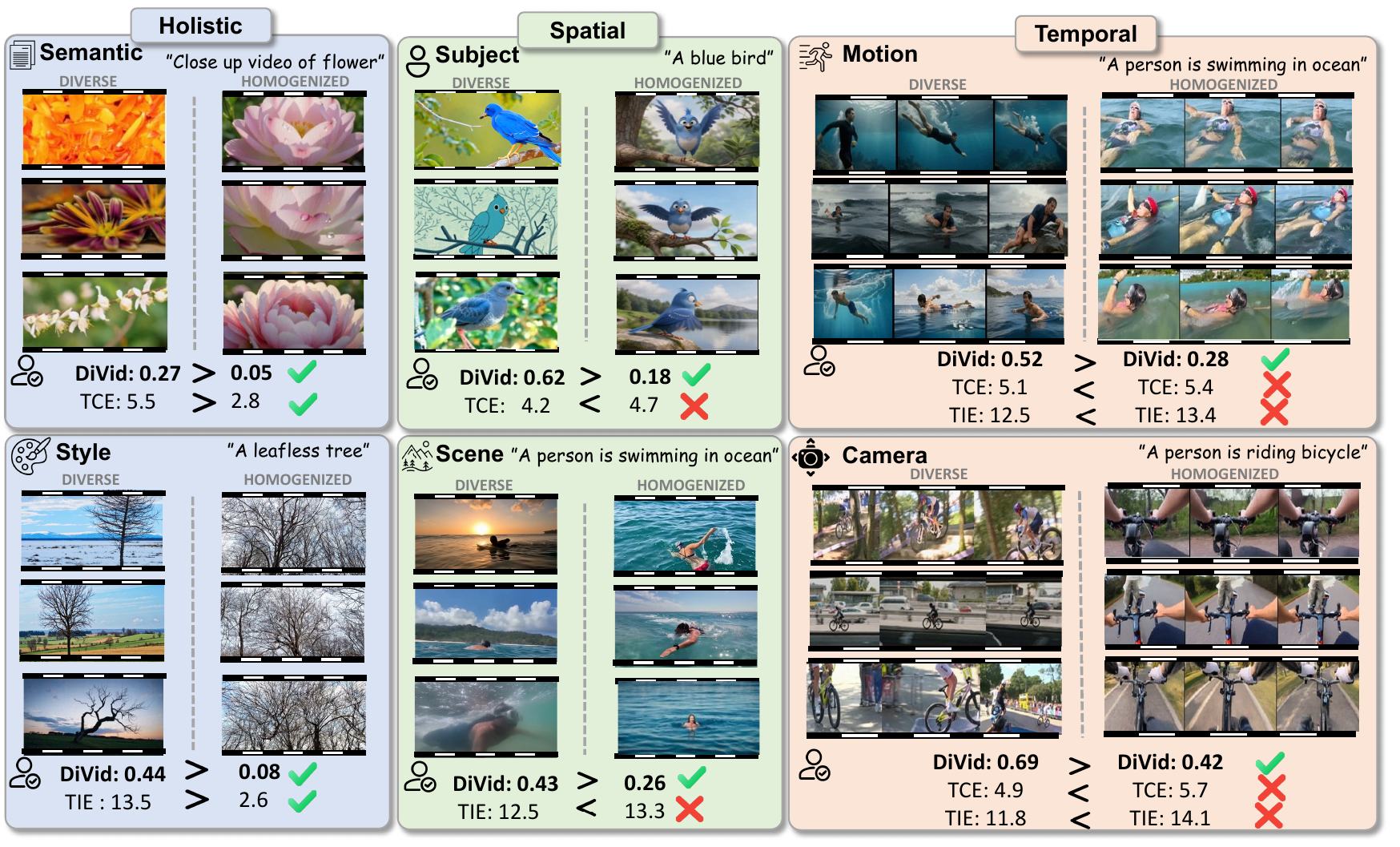}
    \vspace{-10pt}
    \caption{\textbf{Diagnosing dimension-specific homogenization}:
For each dimension, the left set exhibits greater target-factor variation than the right set under the same prompt. DiVid consistently assigns the expected ordering in all dimensions, whereas global metrics such as TCE and TIE can fail to capture factor-level differences.
    }
    \label{fig:intro}
\end{figure*}

Recent advances in video generation have enabled models to produce visually compelling and increasingly realistic videos from text prompts \citep{ho2022video,brooks2024video,hu2026evolution}. However, as these models become more aligned for high-quality generation, a limitation is becoming increasingly apparent: different samples generated from the same prompt often converge to highly similar outputs (illustrated in Figure~\ref{fig:intro}). A model may repeatedly generate the same subject appearance, scene composition, camera trajectory, or motion pattern, despite the large space of possible valid interpretations. Such homogenization has been observed in large language models and text-to-image systems \citep{jiang2025artificialhivemind, han2025poet, dombrowski2025diversity, teotia2025dimcim,zhang2026intricatedance}. Video encodes spatiotemporal signals with far broader creative spaces, but its diversity remains poorly understood.

Diversity is a fundamental capability of generative models rather than merely an aesthetic preference. In practical applications, users often provide high-level intentions instead of fully specified scripts and expect generation systems to explore multiple plausible creative directions \citep{hu2022make, palani2024evolving,davis2024fashioning, choi2026ideablocks}. Limited diversity reduces the usefulness of generation systems by requiring excessive resampling, and may compress human creativity \citep{jiang2025artificialhivemind} and amplify bias \citep{bianchi2023stereotypes,luccioni2023stablebias} at billion-user scale.
More fundamentally, homogenization reveals where models fail to represent the underlying distribution of possible visual worlds \citep{zhu2024sora,jena2025elucidating}: a model that defaults to certain appearances, environments, or motion patterns may reflect biases in training data or objectives \citep{sun2026veat, zhong2026fairt2v, chen2026taming}.
Long-tailed class-imbalanced training data causes model collapse on tail concepts \citep{qin2023cbdm, xu2024rethinking, hong2026improving}, while balanced data resampling degrades overall fidelity \citep{qin2023cbdm, hu2026pdcbdm}. Targeted recovery thus requires locating the collapsed dimension.

Despite its importance, diagnosing diversity in video generation remains challenging. Existing video generation benchmarks primarily focus on quality, alignment and temporal consistency \citep{huang2024vbench, sun2025t2vcompbench, liu2024evalcrafter}.
Beyond individual samples, recent diversity metrics \citep{zheng2025vbench2, kazimi2026dppgrpo} measure it through global scalars, which obscure fundamentally different failure modes. For example, a model may generate diverse scenes while producing identical camera motions, or vary subject appearance while maintaining the same visual style. These models may obtain similar overall diversity scores but require completely different solutions.
Existing metrics thus answer \emph{whether} diversity exists; advancing diversity requires diagnosing \emph{where and how} it collapses.

In this work, we introduce \textbf{DiVid}, a dimension-resolved diagnostic framework for understanding diversity in video generation. Rather than a single global property, DiVid profiles diversity along six complementary and human-aligned dimensions: Semantic and Style diversity at the holistic level, and Subject, Scene, Motion, and Camera diversity at the factor level, each complemented by quality and faithfulness metrics to ensure that measured diversity is genuine. For each dimension, we design a dedicated extraction-and-comparison pipeline built from off-the-shelf computer vision tools, enabling reproducible, localized diagnosis of diversity strengths and failures.

We conduct extensive evaluations on representative video generation models. Our results reveal that diversity is highly dimension-specific: no model dominates across all dimensions, and different models exhibit distinct collapse patterns. Models with strong global diversity scores still collapse on temporal factors such as motion and camera. Lowering the guidance scale does not uniformly lift all dimensions, and diversity rankings disagree with quality and faithfulness rankings. These rankings persist after filtering unfaithful generations, confirming that they reflect genuine capability gaps rather than off-prompt variation. These findings demonstrate that existing global metrics can mask important capability differences between models.

Through controlled prompt interventions that manipulate individual dimensions, we identify two fundamental bottlenecks. First, \emph{default-mode convergence}: when a target factor is unspecified or only broadly hinted, models converge to dominant patterns rather than exploring plausible alternatives. Second, \emph{realization gaps}: explicitly enumerating alternatives expands output span, but models do not consistently realize the requested candidates, particularly for Motion and Camera, echoing the guidance-scale findings above. Improving diversity thus requires dimension-aware objectives and control signals, not a single universal diversity metric.

Our contributions are summarized as follows: 1) We introduce \textbf{DiVid}, a dimension-level diagnostic framework that decomposes video diversity into six interpretable dimensions and provides localized analysis of diversity collapse;
2) We systematically evaluate representative video generation models and reveal that diversity is highly dimension-specific, with substantial gaps hidden by existing global metrics;
3) Through controlled intervention studies, we identify two fundamental diversity bottlenecks, default mode convergence and realization gaps, providing actionable directions for data curation, training objectives, and generation strategies.

\section{Related Work}

\noindent \textbf{Video Generation Evaluation.} Video generation benchmarks have progressed from general-purpose protocols \citep{liu2023fetv, liu2024evalcrafter} to fine-grained dimensions \citep{huang2024vbench, sun2025t2vcompbench, zheng2025vbench2}. However, these efforts predominantly score quality or faithfulness on a per-video basis.
When diversity is considered, some works provide holistic global scores \citep{friedman2023vendi,huang2024vbench,ibarrola2024tietce,zheng2025vbench2}.
Recent prompt-aware methods separate model-induced from prompt-induced variation \citep{ospanov2025scendi,jalali2026conditionalvendi,jalali2026sparke}.
However, they provide limited attribution to the subject, scene, or motion level, and thus lack the fine-grained diagnostic capability.

\noindent \textbf{Diversity and Homogeneity in Generative Models.} Homogenization in generative AI has been independently proved in text \citep{jiang2025artificialhivemind}, image \citep{han2025poet, dombrowski2025diversity}, and video generation \citep{zhong2026fairt2v, sun2026veat}, with alignment training further shown to worsen mode collapse \citep{liu2026diversegrpo, chen2026taming}.
Existing mitigation strategies expand the output distribution through prompt rewriting \citep{kazimi2026dppgrpo, kazimi2026diva}, joint sampling \citep{liu2026cpdv}, or reward optimization \citep{liu2026diversegrpo}, yet predominantly rely on global objectives. While diversity evaluation has been studied in image generation \citep{han2025poet, teotia2025dimcim, astolfi2024paretocdr, zhang2026intricatedance, albuquerque2025attrcondbench}, such systematic evaluation remains largely absent in video generation.

\begin{figure*}[h]
    \centering
    \includegraphics[width=0.9\textwidth]{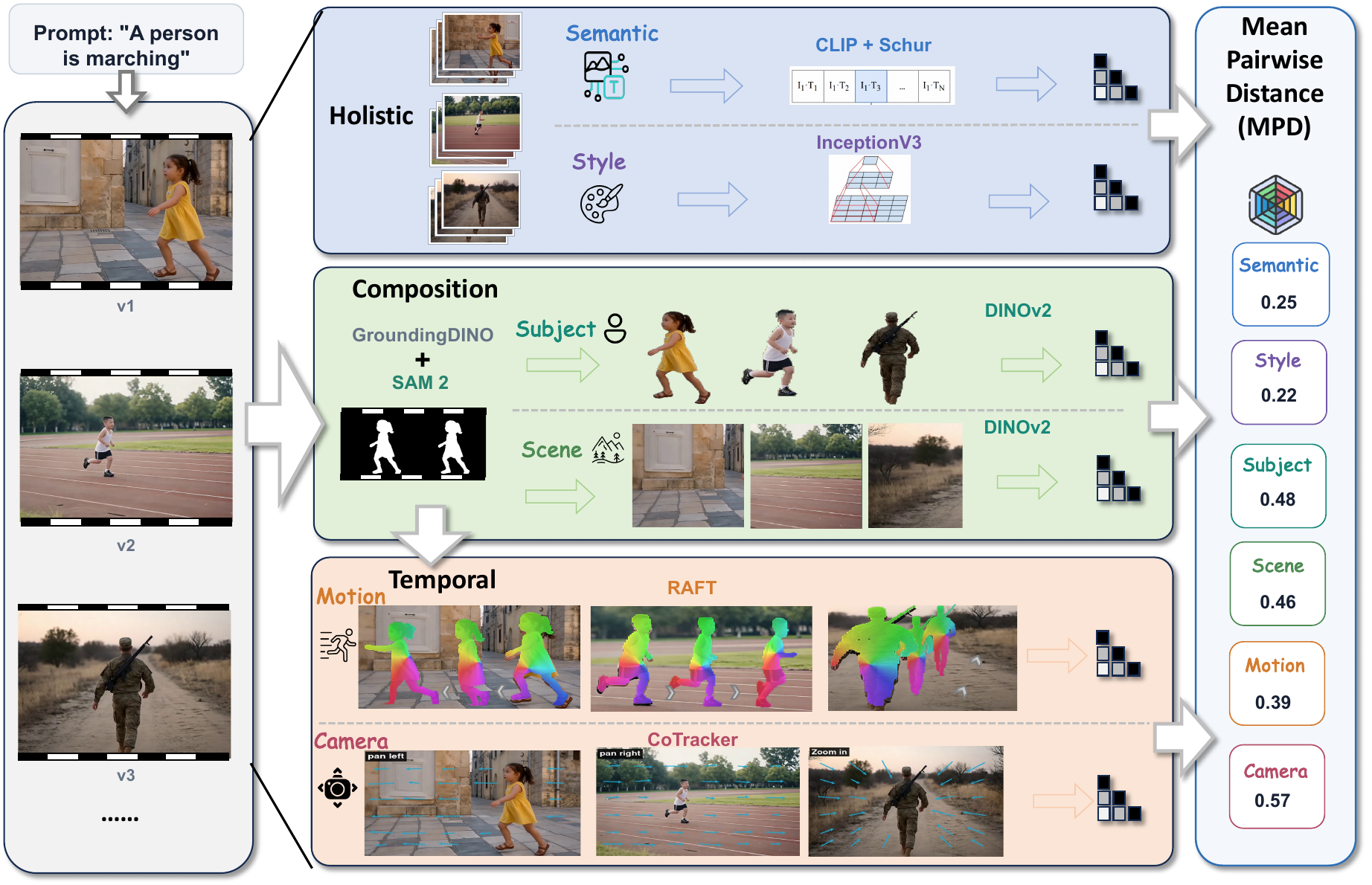}
    \vspace{-8pt}
    \caption{\textbf{Overview of the DiVid Framework}. Given a set of videos generated from a single prompt, our diagnostic framework extracts representations across six spatiotemporal dimensions to compute fine-grained diversity scores.}
    \label{fig:pipeline}
\end{figure*}

\section{DiVid: Framework Design}
\label{sec:benchmark}

\subsection{Definition}

Given a text prompt $p$, a video generation model $m$, and a set of videos $\mathcal{V}_{m,p}=\{v_i\}_{i=1}^{N}$ independently sampled under the same generation settings and different random seeds, we quantify how differently the samples realize prompt-compatible video solutions. Video diversity is a prompt-conditioned, set-level property, fundamentally distinct from single-video quality or instruction faithfulness. However, a video simultaneously encodes 4D spatiotemporal signals. A single holistic embedding can summarize overall dissimilarity but cannot localize where variation originates.

\subsection{Six-Dimensional Diagnostic Pipeline}
\label{sec:pipeline}
Guided by film theory and video representation studies \citep{bordwell2013film, sun2023moso, lin2025camerabench}, a video can be perceived holistically through its content semantics and overall appearance, and be further decomposed along spatial composition (foreground subject, background scene) and temporal dynamics (subject motion, camera movement).
Accordingly, we construct a two-level, six-dimensional diagnostic framework. The Holistic level (Semantic, Style) captures overall variation using established representation spaces; the Factor level further decomposes diversity into four dimensions: spatial composition (Subject, Scene) and temporal dynamics (Motion, Camera).
Figure~\ref{fig:pipeline} illustrates the complete framework pipeline.

\paragraph{Design overview.}
All dimensions share a unified evaluation paradigm: extract a dimension-specific representation for each video, compute a pairwise similarity matrix, and aggregate it into a set-level diversity score. The holistic Semantic and Style dimensions adapt image-level diversity kernels \citep{ospanov2025scendi, ibarrola2024tietce} to video through frame-level aggregation; the remaining four factor-resolved dimensions introduce pipelines combining detection, segmentation, tracking, and motion decomposition to achieve dimension-level decoupled evaluation.

\paragraph{Holistic level.}
The \textbf{Semantic} dimension adopts the Schur-complement decomposition of the CLIP \citep{radford2021clip} kernel proposed by Scendi \citep{ospanov2025scendi}, subtracting the prompt-alignment component to capture semantic differences after removing prompt-shared information.
The \textbf{Style} dimension aggregates frame-level InceptionV3 \citep{szegedy2016rethinking} representations to capture differences in overall appearance, color, lighting, texture, style, and layout.

\paragraph{Factor level: spatial composition.} Subject and Scene rely on a subject localization pipeline: GroundingDINO \citep{liu2024groundingdino} detects the subject in a frame and SAM~2 \citep{ravi2025sam2} propagates temporally consistent masks across frames.
The \textbf{Subject} dimension extracts subject crops and encodes them with DINOv2 \citep{oquab2024dinov2} to measure differences in subject appearance.
The \textbf{Scene} dimension masks the subject region and encodes the resulting background-focused frames with DINOv2 to capture environmental differences.

\paragraph{Factor level: temporal dynamics.} This group disentangles two motion components: global camera movement and subject motion.
The \textbf{Camera} dimension mainly fits a global affine transformation (translation, scaling, and rotation) to background trajectories extracted by CoTracker \citep{karaev2024cotracker}, capturing differences in camera movement patterns and pace.
The \textbf{Motion} dimension first compensates subject trajectories by removing estimated camera transform, then combines the resulting relative trajectories with canonicalized RAFT \citep{teed2020raft} optical flow to capture subject locomotion and internal deformation.
This joint design minimizes leakage between camera movement and subject motion.

\paragraph{Metric Aggregation.} Each dimension ultimately yields an $N \times N$ similarity matrix, which is aggregated using Mean Pairwise Distance (MPD) and Vendi Score \citep{friedman2023vendi}. MPD serves as our primary metric for its interpretability and robustness to $N$, where larger values indicate more variation in the corresponding representation space. Vendi Score provides a complementary estimate of the effective number of distinct modes via eigenvalue entropy.

\subsection{Metric Reliability}
\label{sec:reliability}

\paragraph{Human Evaluation.}
For diversity dimensions and faithfulness, we design separate blind human evaluation protocols. All six metrics correlate significantly with human judgments of set-level diversity (Spearman's $\rho=0.531$--$0.604$, Kendall $\tau_b=0.471$--$0.539$, all $p<0.001$).
Factor-level faithfulness metrics achieve $85.4\%$--$90.3\%$ directional agreement with human ratings.
For all Diversity and Faithfulness metrics, the human evaluation agreement (Krippendorff's $\alpha$) is greater than 0.667. These results support the perceptual validity of our diagnostic framework.
Full protocols and statistics are in the Appendix \ref{app:human-evaluation}.

\paragraph{Framework Reliability.}
We validate the shared localization and each dimension through automatic validity filtering, human mask inspection, and strict-validity sensitivity analysis. The prompt--dimension invalid rates are only 0.83\% for Subject and Scene, 0.76\% for Motion, and 0.69\% for Camera; invalid records are excluded rather than assigned zero. The corresponding four-dimension model-level rate ranges from 0\% to 2.79\%. Further human verification on 300 randomly sampled masks indicates that 93.3\% contain neither salient foreground omission nor background leakage. Moreover, the dimension leaders remain unchanged when every model is required to retain at least 8 of 10 valid videos. Details and sensitivity checks are in Appendix~\ref{app:failure}.

\paragraph{Diagnostic Utility.}
We prove the non-interchangeability and diagnostic value of the four factor-level dimensions through prompt-grouped cross-factor prediction, leave-one-factor-out diagnosis, and dimension-selective pair retrieval. The other three factors leave 58.1--73.4\% of each held-out factor's variance unexplained; removing one factor changes the prompt-level top-model set for 19.5--26.7\% of prompts. Each factor retrieves selective video pairs in 74.7--93.3\% of the prompts. Thus, Subject, Scene, Motion, and Camera expose factor-specific, decision-relevant failures that a global metric would miss, detailed in Appendix~\ref{app:diagnostic-utility}.

\begin{table*}[t]
\centering
\resizebox{\textwidth}{!}{
\begin{tabular}{l rrrrr rr rr rrrr}
\toprule
& \multicolumn{5}{c}{\textbf{Faith / Quality}}
& \multicolumn{2}{c}{\textbf{Prior Metrics}}
& \multicolumn{2}{c}{\textbf{Holistic}}
& \multicolumn{4}{c}{\textbf{Factor Level}} \\
\cmidrule(lr){2-6}
\cmidrule(lr){7-8}
\cmidrule(lr){9-10}
\cmidrule(lr){11-14}
Model & IF & AQ & IQ & TF & MS & TCE & TIE & Semantic & Style & Subject & Scene & Motion & Camera \\
\midrule
Wan2.2-5B (g=5)
& 0.848 & 0.543 & 0.672 & \textbf{0.980} & 0.989 & \textbf{5.29} & \textbf{13.36} & \textbf{0.169} & \textbf{0.234} & \textbf{0.447} & \textbf{0.426} & 0.326 & 0.097 \\
Wan2.2-14B
& 0.915 & \textbf{0.614} & 0.699 & 0.972 & 0.983 & 4.54 & \underline{13.29} & 0.144 & \underline{0.224} & 0.412 & \underline{0.394} & 0.344 & 0.148 \\
CogVideo
& 0.839 & 0.516 & 0.622 & 0.973 & 0.985 & 4.12 & 12.66 & 0.142 & 0.214 & \underline{0.414} & 0.377 & 0.323 & 0.045 \\
Hunyuan
& 0.893 & 0.541 & 0.662 & 0.965 & \textbf{0.991} & 5.05 & 13.07 & 0.142 & 0.197 & 0.392 & 0.378 & \textbf{0.433} & \textbf{0.490} \\
\midrule
Wan2.7
& \underline{0.963} & 0.563 & \underline{0.722} & 0.972 & 0.985 & 4.94 & 12.74 & 0.158 & 0.201 & 0.397 & 0.390 & 0.413 & 0.303 \\
HappyHorse
& \textbf{0.986} & 0.594 & \textbf{0.729} & \underline{0.978} & \textbf{0.991} & 2.77 & 11.45 & 0.093 & 0.123 & 0.250 & 0.223 & 0.326 & 0.193 \\
Seedance
& 0.955 & \underline{0.595} & 0.667 & 0.970 & 0.990 & \underline{5.16} & 13.09 & \underline{0.160} & 0.162 & 0.380 & 0.364 & \underline{0.425} & \underline{0.463} \\

\specialrule{0.10em}{0.45em}{0.35em}

Wan2.2-5B (g=3)
& 0.849 & 0.534 & 0.670 & 0.980 & 0.990 & 5.63 & 13.83 & 0.180 & 0.248 & 0.479 & 0.463 & 0.153 & 0.050 \\
Wan2.2-5B (g=7)
& 0.875 & 0.549 & 0.680 & 0.979 & 0.989 & 4.79 & 13.22 & 0.147 & 0.229 & 0.428 & 0.402 & 0.141 & 0.048 \\
\bottomrule
\end{tabular}%
}
\caption{
Faithfulness, quality and diversity profiles pooled over the 206 general prompts. The table is divided into open-source models, closed-source models, and two ablations. Results are macro-averaged over available prompt-level measurements. \textbf{Bold} and \underline{underlined} denote the highest and second-highest values for normal setting.
}
\label{tab:main_results}

\end{table*}

\subsection{Additional Metrics}
\paragraph{Instruction Faithfulness (IF).}
It measures whether a video adheres to the prompt, ensuring valid diversity.
A VLM (Qwen3.6-27B \citep{bai2025qwen3}) judge extracts the constraints for a single factor-level dimension (subject, scene, motion, or camera) from the prompt and independently assesses whether the generated video satisfies that dimension, assigning a score from 1 to 5, which is then linearly normalized to [0.2, 1]. The unconstrained dimension is marked N/A and Overall IF averages all applicable factor scores.

\paragraph{Quality.} We employ Aesthetic Quality (AQ) and Imaging Quality (IQ) from VBench \citep{huang2024vbench} to characterize visual fidelity, while the original VBench Temporal Flickering (TF) and Motion Smoothness (MS) metrics assess motion quality, ensuring that the observed diversity does not arise from artifacts or temporal degradation.

\paragraph{Reference diversity evaluation baselines.} Following DPP-GRPO \citep{kazimi2026dppgrpo}, we report TCE and TIE as global set-level references that measure the spread of generated videos in CLIP and Inception feature spaces, respectively.

\section{Main Evaluation}
\label{sec:main-eval}

\subsection{Experimental Setup}
\label{sec:setup}
\textbf{Prompt Construction.} We construct our prompt set from two existing benchmarks to ensure generality, reproducibility, and broad coverage: VBench-Category \citep{huang2024vbench} (8 general content categories), VBench-Dimension (10 quality dimensions), and T2V-CompBench \citep{sun2025t2vcompbench} (7 categories of compositional constraints), sampling 10, 7, and 8 prompts per category respectively, for a total of 206 general prompts. All results reported are aggregated across the full prompt set; per-source breakdowns are provided in Appendix~\ref{app:prompt-transfer} and show consistent trends.

\paragraph{Tested Models.}  Open-source models include Wan2.2-5B \citep{wang2025wan}, Wan2.2-14B \citep{wang2025wan}, CogVideoX1.5-5B \citep{yang2025cogvideox}, HunyuanVideo-1.5 \citep{wu2025hunyuanvideo}. Closed-source models include Wan2.7 \citep{wan2.7}, HappyHorse-1.5 \citep{happyhorse1.5}, and Seedance-1.5-pro \citep{seedance2025seedance}.

\paragraph{Evaluation Setup.}  For each prompt, each model independently generates 10 videos using distinct random seeds. All videos are generated at 5 seconds with a 16:9 aspect ratio, guidance scale 5.0 for open-source models. For Wan2.2-5B, guidance scales of 3.0 and 7.0 are evaluated for ablation. Other parameters are set to default, prompt extension disabled.
All model-level statistics are computed with the prompt as the unit of analysis.

\subsection{Dimension-Specific Diversity}
\label{sec:main-results}

As shown in Table~\ref{tab:main_results}, homogenization in video generation is dimension-specific and cannot be characterized by any single global score.

\begin{figure}[t]
    \centering
    \includegraphics[width=0.9\linewidth]{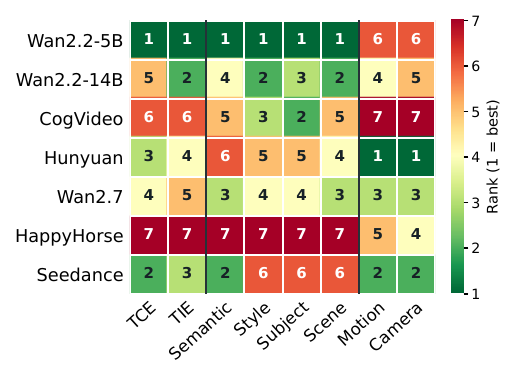}
    \caption{Rank Heatmap on all diversity metrics.}
    \label{fig:heat_map}
\end{figure}

\begin{figure*}[t]
    \centering
    \includegraphics[width=0.8\linewidth]{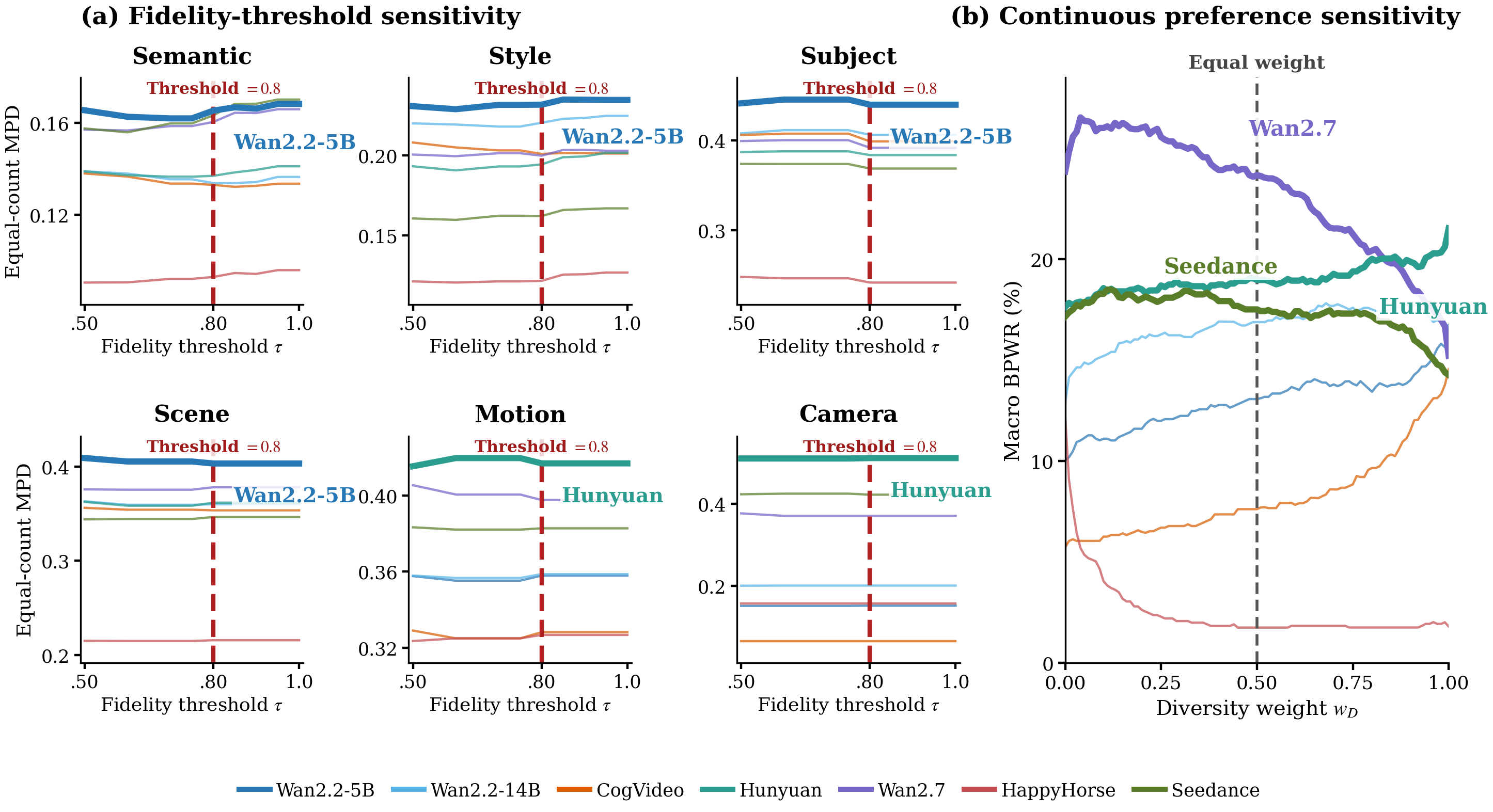}
    \vspace{-8pt}
    \caption{Faithfulness--Diversity trade-off. (a) Equal-count MPD ($k=3$) under faithfulness threshold $\tau$; colored curves highlight the Top-1 model at $\tau=0.8$. (b) Six-dimension macro BPWR under diversity weight $w_D$.}
    \label{fig:tradeoff}
\end{figure*}

\paragraph{Diversity Ranks Reverse Across 6 Dimensions.}
In Figure~\ref{fig:heat_map}, Wan2.2-5B ranks first in Semantic, Style, Subject, and Scene, but falls into the bottom two in Motion and Camera. Hunyuan exhibits the opposite profile: it ranks first in both Motion and Camera, but only fourth to sixth across the four content dimensions.
Similar reversals appear beyond the two leaders. Seedance ranks second in both temporal dimensions but sixth in Style, Subject, and Scene, whereas CogVideo ranks second in Subject but last in Motion and Camera. Overall, six of the seven models span at least three rank positions across the six dimensions.
Prompt-paired bootstrap (20,000 resamples) and two-sided Wilcoxon signed-rank tests with Benjamini--Hochberg correction confirm the central rank reversal, detailed in Appendix \ref{app:complete-results-final}.

\paragraph{Global Metrics Conflate Opposite Diversity Collapse.}
Both TCE and TIE rank Wan2.2-5B first. These scores are close to Hunyuan's 5.05 and 13.07, yet their factor-level results are fundamentally different: Hunyuan achieves $1.33\times$ higher Motion MPD and $5.05\times$ higher Camera MPD, while Wan2.2-5B leads all four content dimensions.
Additionally, Camera MPD spans a 10.8$\times$ gap---from 0.490 (Hunyuan) to 0.045 (CogVideo).
Thus, global metrics can summarize aggregate spread, but they neither preserve the full factor-level profile nor identify the source of homogenization. This diagnostic gap motivates factor-specific evaluation and optimization objectives.

\paragraph{Effect of CFG Scale.}
Lowering CFG from $g=5$ to $g=3$ increases Semantic, Style, Subject, and Scene diversity by 5.8--8.7\%. Raising CFG to $g=7$ improves IF, AQ, and IQ, respectively.
However, both deviations from $g=5$ substantially reduce Motion and Camera diversity, indicating that CFG redistributes exploration across factors rather than uniformly controlling diversity. Motion and Camera require further control signals.

\subsection{Joint Analysis on Quality, Faithfulness, and Diversity}
\label{sec:tradeoff}

No evaluated model performs strongly across faithfulness, quality, and all diversity dimensions. We analyze the three jointly: where they misalign, whether the measured diversity is genuine, and which model to choose under a stated preference.

\paragraph{Misalignment on Quality, Faithfulness and Diversity.} Achieving high quality, instruction faithfulness, and rich diversity simultaneously remains a fundamental challenge in current video generation. HappyHorse ranks first in instruction faithfulness (IF=0.986) and imaging quality (IQ=0.729), yet ranks last in Semantic, Style, Subject, and Scene diversity. Conversely, Wan2.2-5B leads the four dimensions. At the prompt level, 35 of 49 significant correlations (after FDR correction) are negative (Table \ref{tab:app4-prompt-correlations}). Selecting a model from a quality leaderboard may favor the homogeneous one.

\begin{table*}[t]
\centering
\resizebox{\textwidth}{!}{%
\begin{tabular}{l|cccc|ccc|cccc}
\toprule
\multirow{2}{*}{Model}
& \multicolumn{4}{c|}{\textbf{Hinted}}
& \multicolumn{3}{c|}{\textbf{Omitted}}
& \multicolumn{4}{c}{\textbf{Enumerated}} \\
\cmidrule(lr){2-5}
\cmidrule(lr){6-8}
\cmidrule(lr){9-12}
& Subject & Scene & Motion & Camera
& Scene & Motion & Camera
& Subject & Scene & Motion & Camera \\
\midrule
Wan2.2-5B 
& +0.237 & +0.014 & \textbf{+0.128} & +0.032
& +0.041 & +0.049 & +0.003
& +0.450 & +0.395 & +0.168 & +0.031 \\
Wan2.2-14B
& +0.203 & +0.028 & +0.100 & +0.219
& +0.034 & -0.021 & +0.009
& +0.461 & \textbf{+0.444} & +0.174 & +0.167 \\
CogVideo
& +0.138 & +0.056 & +0.034 & +0.010
& +0.041 & +0.010 & +0.004
& +0.411 & +0.397 & +0.095 & +0.017 \\
Hunyuan
& +0.209 & +0.075 & -0.047 & +0.081
& +0.076 & -0.020 & +0.093
& +0.492 & +0.308 & +0.067 & +0.281 \\
Wan2.7
& +0.234 & \textbf{+0.109} & +0.067 & +0.398
& \textbf{+0.105} & -0.004 & +0.109
& +0.468 & +0.341 & +0.146 & +0.534 \\
HappyHorse
& +0.091 & +0.030 & +0.105 & +0.295
& -0.015 & +0.015 & +0.085
& \textbf{+0.566} & +0.397 & \textbf{+0.212} & +0.606 \\
Seedance
& \textbf{+0.269} & +0.080 & +0.095 & \textbf{+0.433}
& +0.076 & \textbf{+0.049} & \textbf{+0.340}
& +0.460 & +0.305 & +0.159 & \textbf{+0.614} \\
\midrule
\textbf{Mean}
& +0.197 & +0.056 & +0.069 & +0.210
& +0.051 & +0.011 & +0.092
& +0.473 & +0.370 & +0.146 & +0.321 \\
\textbf{Positive Models}
& 7/7 & 7/7 & 6/7 & 7/7
& 6/7 & 4/7 & 7/7
& 7/7 & 7/7 & 7/7 & 7/7 \\
\bottomrule
\end{tabular}%
}
\vspace{-8pt}

\caption{Mean MPD changes relative to the Specified condition; \textbf{bold} denotes the largest responses within each column, while Positive models reports the number of models with positive mean changes.
}
\label{tab:explore}
\end{table*}

\begin{table}[h]
    \centering
    \vspace{-6pt}
    \scriptsize
    \setlength{\tabcolsep}{3pt}
    \renewcommand{\arraystretch}{1.15}
    \parbox{\linewidth}{
    \textbf{Specified:}
    \emph{A golden retriever is resting on a beach, static camera.}
    }
    \vspace{2pt}
    \begin{tabularx}{\linewidth}{@{}l l l X@{}}
    \toprule
    Dimension & Omitted & Hinted & Enumerated \\
    \midrule
    Subject & N/A & \emph{an animal}
            & \{dog, cat, horse, \ldots\} \\
    Scene   & remove \emph{on a beach} & \emph{somewhere}
            & \{beach, forest, city, \ldots\} \\
    Motion  & remove \emph{resting} & \emph{in motion}
            & \{walks, runs, jumps, \ldots\} \\
    Camera  & remove \emph{static camera} & \emph{camera moves}
            & \{pan, zoom, dolly, \ldots\} \\
    \bottomrule
    \end{tabularx}
    \caption{Illustration of the prompt settings.}
    \label{tab:prompt_settings}
\end{table}

\paragraph{Validity of Measured Diversity.} To test whether high MPD is a valid measurement or caused by off-prompt samples, we retain three faithful videos per model and prompt with faithfulness $\geq 0.8$, which means they closely follow the prompt (Figure~\ref{fig:tradeoff}a). On the jointly valid prompt sets, dimension-level rankings remain unchanged: Wan2.2-5B still leads the four dimensions while retaining more than 98.3\% diversity of its original MPD, and Hunyuan leads in Motion and Camera. The dimension-specific rankings therefore cannot be explained by unfaithful outputs or unequal pair counts.
For quality, all models attain high TF and MS. Especially, Hunyuan ranks first in motion, camera and MS, so its diversity cannot be largely explained by artifacts or degraded motion smoothness.

\paragraph{ Preference-Conditioned Model Selection. } Sweeping the diversity weight $w_D$ (Figure~\ref{fig:tradeoff}b) reveals no universally optimal model: Wan2.7 ranks first for $w_D\in[0,0.83]$; Hunyuan takes over at $w_D\geq0.84$. The overall recommendation shifts from Wan2.7 to Hunyuan as diversity receives greater weight. The best fidelity--diversity compromise depends on which dimensions matter most. The six-dimension macro curve averages win rates, not raw MPD, and serves as a rough recommendation summary. Details are in Appendix \ref{app:Faithfulness-Diversity}.

\section{Locating Diversity Bottlenecks via Prompt Intervention}
\label{sec:5_explore}

Section~\ref{sec:main-eval} reports dimension-wise diversity profiles on general-purpose benchmark prompts.
Here, we use controlled prompts that deliberately leave a target dimension open, isolating scenarios where diversity in the target dimension is expected. This design distinguishes three behavioral regimes: default convergence under open prompts, diversity activation via generic cues, and candidate realization through explicit enumeration.

\subsection{Experimental Setup}
\paragraph{Prompt Construction.} We use a Specified prompt that specifies all four dimensions as the paired baseline, then vary only one target dimension under three interventions with progressively increasing information (Table~\ref{tab:prompt_settings}). \textbf{Specified} prompts ensure dimensional independence as much as possible: replacing one dimension does not constrain the others. \textbf{Omitted} removes the target-dimension description, exposing the model's unconstrained prior. \textbf{Hinted} replaces it with a dimension-level generic cue, testing whether a broad prompt can activate the dimension. \textbf{Enumerated} uses an LLM to enumerate prompts differing only in the target dimension, measuring model's controllability to realize predefined variation.
All interventions share the same semantic core, distinguishing among settings with no cue, a broad cue, and explicit candidates. The experiment covers 20 semantic groups. Annotation details are in Appendix \ref{app5:prompt-design}.

\paragraph{Evaluation Metrics.} For each model, semantic group, and target dimension, we compute the paired difference between the intervention MPD and the Specified MPD, denoted as $\Delta$. Model-level results are the mean $\Delta$ over the 20 semantic groups. We also report directional consistency among models: 7/7 model agreement is robust, 6/7 is a strong trend.\footnote{MPD metrics across dimensions are in different representation spaces and are not directly comparable.}

\begin{figure}[h]
    \centering
    \includegraphics[width=0.95\linewidth]{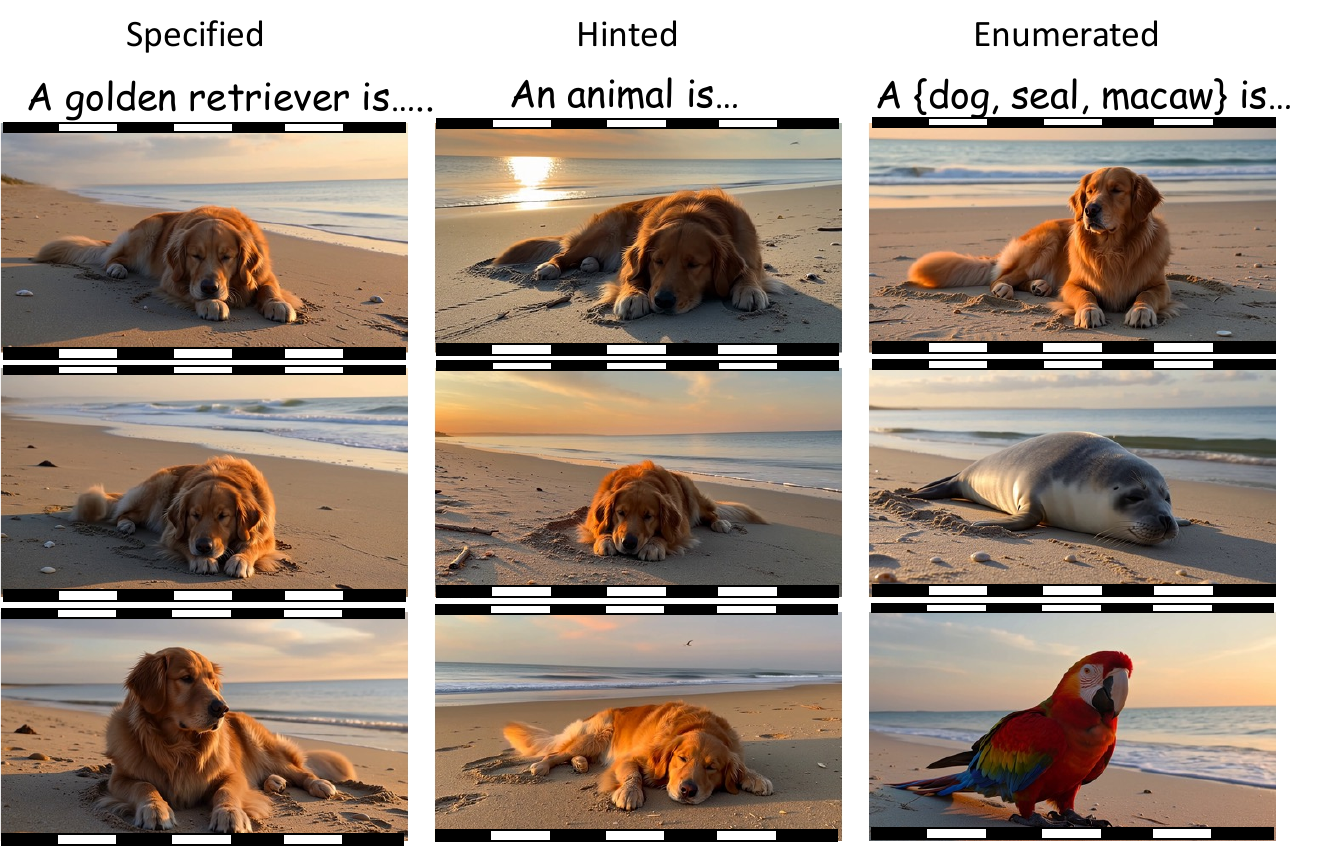}
    \caption{A subject convergence case: The model defaults ``an animal'' to ``a golden retriever''.}
    \label{fig:subject-collapse}
\end{figure}

\subsection{Default Mode Convergence}
\paragraph{Open Prompts Trigger Default Preference.} Table \ref{tab:explore} shows that, although removing or explicitly relaxing a constraint is expected to increase diversity, every dimension still contains cases that revert to a default preference. For example, under Omitted, Motion improves for only four models, with a mean change of merely $+0.011$.
Changing the generic wording gets consistent results: \texttt{in motion} and \texttt{performing a clear action} yield Motion improvements of $+0.069$ and $+0.054$, respectively, both with positive responses in $6/7$ models; similarly, \texttt{camera moves} and \texttt{with clear camera movement} yield Camera improvements of $+0.210$ and $+0.213$, respectively, both with $7/7$ model agreement.
For Scene dimension, after adding a generic cue, \texttt{somewhere} yields only a $+0.005$ improvement over Omitted.  These results indicate that open constraints primarily activate high-probability default solutions rather than reflecting the natural variance of the open prompt.

\paragraph{Dimension-dependent Prompt Sensitivity.}
The response to prompt specificity differs by dimension: For Scene dimension, the generic cue \texttt{somewhere} improves over Omitted by only $+0.005$ in the mean row, whereas explicit enumeration lifts diversity by $+0.313$. The temporal dimensions instead respond gradually: \texttt{in motion} and \texttt{camera moves} improve over Omitted by $+0.058$ and $+0.118$, and Motion enumeration further gains $+0.077$. These profiles suggest that discrete, categorical dimensions such as Subject and Scene are unlocked only by explicit candidate enumeration, whereas temporal dimensions benefit from coupling high-level candidate planning with low-level execution constraints on trajectory and camera.

\subsection{Realization Gap}
\label{sec:realization_gap}

\begin{figure}[h]
    \centering
\includegraphics[width=1.0\linewidth]{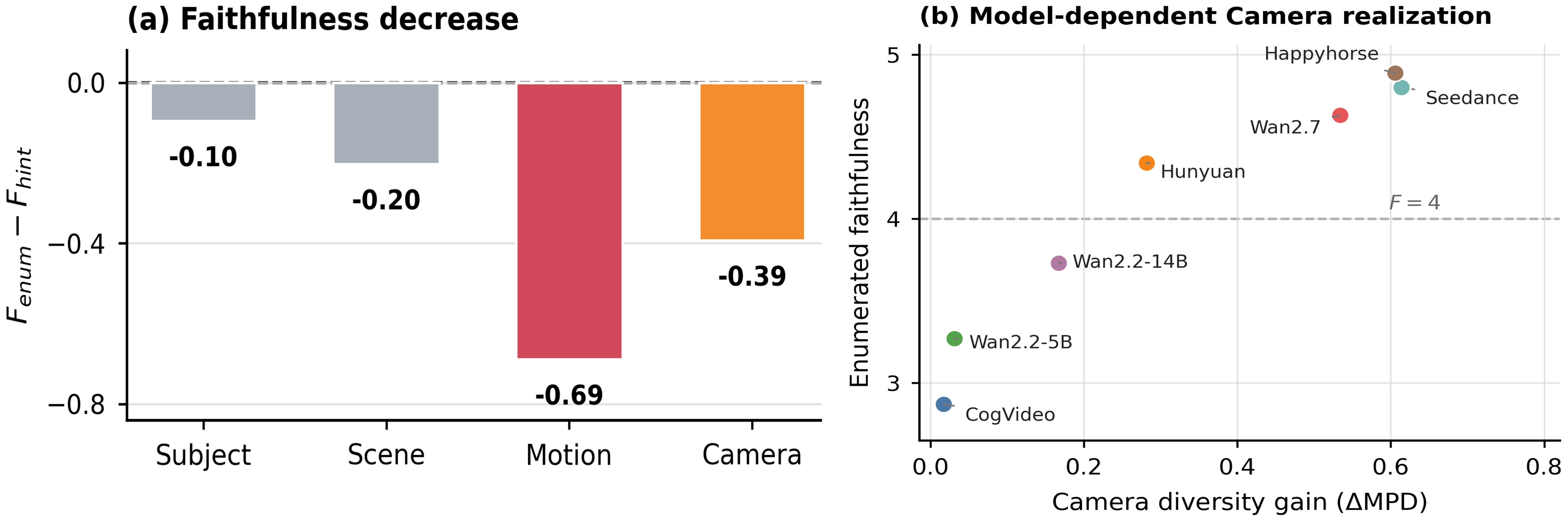}
    \vspace{-8pt}
    \caption{Faithful realization under enumeration on the original faithfulness scale.
    (a) Changes in target faithfulness from Hinted to Enumerated.
    (b) Camera diversity gain versus Enumerated faithfulness.}
    \label{fig:realization}
\end{figure}

\paragraph{Faithful Execution Bottleneck.}
Explicit enumeration consistently improves diversity across all four dimensions for all models. However, compared with the corresponding Hinted prompts, mean Faithfulness decreases in all dimensions, with larger decreases in Motion and Camera than in Subject and Scene (Figure \ref{fig:realization}a).
Thus, larger MPD indicates candidate influence rather than faithful coverage; diversity and execution must be optimized jointly. Consequently, relying solely on text prompts to unlock fine-grained motion diversity is fundamentally bottlenecked, highlighting the necessity for future trajectory-based or multi-modal control signals to resolve temporal homogenization.

\paragraph{Model-dependent Responsiveness.}
Although enumeration obtains positive responses from all models, models differ substantially in realization faithfulness. In Figure \ref{fig:realization}b, Seedance, HappyHorse, and Wan2.7 all show large Camera gains, with high faithfulness, whereas CogVideo and Wan2.2-5B gain slightly with low faithfulness; the latter two also rank last on the main Camera Table~\ref{tab:main_results}.
Furthermore, HappyHorse ranks last in Subject/Scene diversity while faithfully realizing the enumerated Subject and Scene candidates.
The prompt-dependent default preference can likely be mitigated by prompt rewriting or decoding strategies, whereas capacity-limited failures call for better training data or architectural improvements. Our benchmark provides this model-level diagnosis.

\section{Conclusion}
We introduce \textbf{DiVid}, a reproducible dimension-level diagnostic framework that localizes diversity collapse in video generation across six interpretable dimensions.
Systematic evaluations of representative models reveal distinct collapse patterns hidden by global metrics.
Joint analysis on faithfulness and quality confirms that these differences reflect genuine generation capabilities rather than off-prompt variation.
Beyond diagnosis, we find that CFG cannot uniformly control all dimensions, particularly temporal ones, open-ended prompts lead models toward default preferences, and explicitly requested alternatives remain difficult to realize faithfully, again most notably for Motion and Camera.
DiVid helps identify promising future directions: dimension-aware training objectives and richer control signals beyond text prompts for achieving diverse, controllable and faithful video generation.

\subsection*{Ethics Statement}
This research was conducted in strict adherence to the Code of Ethics.
For human evaluation, the annotators we recruited possess a high level of education. They were fairly compensated for their time and effort in evaluation according to criteria.
Our benchmark evaluates videos generated by publicly available T2V models using prompts from open-source benchmarks. Generated content may occasionally contain biased but unharmful content.

{
    \small
    \bibliographystyle{ieeenat_fullname}
    \bibliography{refer}
}

\clearpage
\appendix

\section{Discussion and Future Directions}

\subsection{Dimension Extraction}
The four factor dimensions (Subject, Scene, Motion, Camera) depend on automatic extraction of video constituents. Accurate extraction is an independent line of work: subject detection and segmentation \citep{liu2024groundingdino, ravi2025sam2}, point tracking and optical flow \citep{karaev2024cotracker, teed2020raft}, and camera-motion understanding and pose estimation for video \citep{lin2025camerabench, li2025megasam}. Recovering motion and camera information from 2D frames remains the hardest part \citep{lin2025camerabench, he2025cameractrl}. Our aim is not more accurate extraction, but natural disentanglement of video constituents and set-level measurement of diversity.
Every score is a pairwise similarity computed within the videos of the same prompt, aggregated at the prompt level, and used for relative ranking across models; Relative discriminability, reproducibility, and efficiency are what matter here.
Appendix~\ref{app:failure} establishes the validity and robustness of the metrics through explicit validity controls, comprehensive data-quality statistics, human verification, and sensitivity analyses.

Our extraction accordingly follows the background-compensation paradigm validated in action recognition \citep{wang2013action}, adapted to set-level measurement: we exclude the foreground with a dilated mask and fit a global partial-affine transform (translation, rotation, uniform scale) by RANSAC, which is richer than subtracting an averaged background vector from an averaged foreground vector \citep{sun2025t2vcompbench}; Motion adds a canonicalized deformation channel with two interpretable kernels for relative displacement and internal deformation; Camera aggregates translation, scale, and rotation into a 16-D descriptor;
All key parameters are carefully selected by pilot analyses to reflect its intended factor and are then fixed across all models and prompts.
The pipeline makes dimension-resolved diversity measurement efficient at benchmark scale. As motion and camera estimation for generated video improves in reliability and efficiency, we can update the corresponding modules.

\subsection{Toward Dimension-Aware Diversity Enhancement}
DiVid points to three complementary directions for diversity enhancement: data and training objectives determine which variations a model learns, control interfaces determine how users specify the desired variation, and inference strategies determine how to realize it.

\begin{itemize}[leftmargin=*, itemsep=0pt, topsep=0pt]
\item \textbf{Data and Objectives.}
Section~\ref{sec:5_explore} shows that models often converge to dominant defaults. For example, HappyHorse repeatedly interprets \emph{an animal} as a golden retriever, whereas explicitly listing candidates expands its Subject diversity. Meanwhile, the real-video retrieval reference in Appendix~\ref{sec:real_video_reference} leads in the four content-related dimensions but ranks only fourth in Motion and Camera, suggesting that broad web-video coverage alone does not guarantee equally rich temporal variation. Prior image-generation studies similarly show that class imbalance causes tail-mode collapse, while global rebalancing can reduce overall fidelity \citep{qin2023cbdm,hu2026pdcbdm}. Future methods should therefore curate data based on the diagnosed collapse dimensions and develop factor-specific training objectives that improve coverage and instruction following, rather than uniformly rebalancing the training distribution.
\item \textbf{Control Interfaces.}
Prompt interventions show that different dimensions require different forms of control. Explicitly listing candidates effectively expands discrete content factors, but produces less reliable gains in Motion and Camera. Text alone is therefore insufficient for precise control. Future systems could adopt hierarchical control, using text-level candidate specification for content factors and structured conditions for Motion and Camera, such as trajectories, keypoints, and camera paths.

\item \textbf{Inference and Sampling.}
In Section~\ref{sec:main-results}, we systematically examine how CFG scale affects each diversity dimension.
As a preliminary feasibility check, we try adapting three diversity-enhancement methods from image generation to video: a bell-shaped CFG schedule; frequency-decoupled guidance with $g{=}1.5$ for low frequencies and $g{=}5.0$ for high frequencies \citep{sadat2025guidance}, and condition-annealed sampling that adds Gaussian noise to the T5 text embedding early in sampling and gradually reduces it to zero \citep{sadat2024cads}. In pilot experiments, these methods do not produce consistent improvements across dimensions: some apply only to specific models, while others may trade quality for diversity.
This exposes a common limitation of scalar guidance: it cannot select which factor to diversify. Future inference methods should instead use dimension-specific guidance and factor-aware sampling to explore the target dimension while jointly preserving quality and faithfulness.
\end{itemize}

These three directions share one conclusion: a global diversity metric cannot provide a concrete optimization target, making dimension-level diagnosis essential for targeted enhancement.
Recent controllable video-generation methods use structured camera poses, trajectories, or flow to control camera and object motion \citep{he2025cameractrl,wang2024motionctrl,geng2025motionprompting,burgert2025gowithflow,wang2025ati,feng2025i2vcontrol}, with newer approaches further disentangling these factors or composing their control over longer horizons \citep{meng2026orthomotion,wang2026directing}.
These advances provide practical mechanisms for acting on our diagnosis. DiVid complements them by locating the collapsed factor, evaluating whether a control method expands its intended dimension, and providing a low-cost automatic signal for training or sampling while jointly monitoring quality and faithfulness.

\section{Additional Details of the Framework}
\label{app:benchmark-details}

This appendix provides robustness of generation number, implementation details, validity checks, and scope conditions that complement Section~\ref{sec:benchmark}.

\subsection{Robustness of Generation Number}
\begin{figure*}[h]
\centering
\begin{minipage}[t]{0.49\textwidth}
\centering
\includegraphics[width=\linewidth]{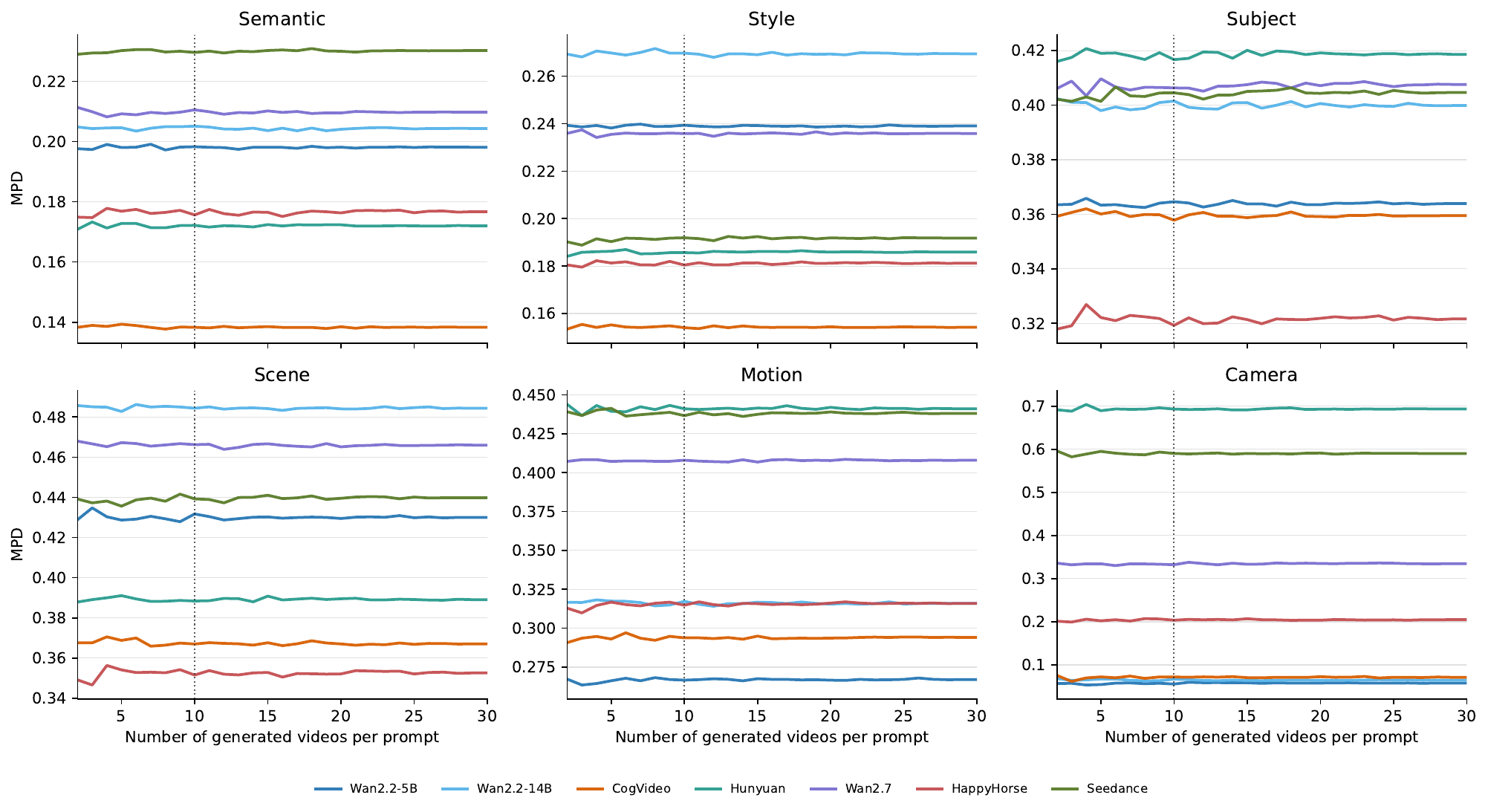}
\end{minipage}\hfill
\begin{minipage}[t]{0.49\textwidth}
\centering
\includegraphics[width=\linewidth]{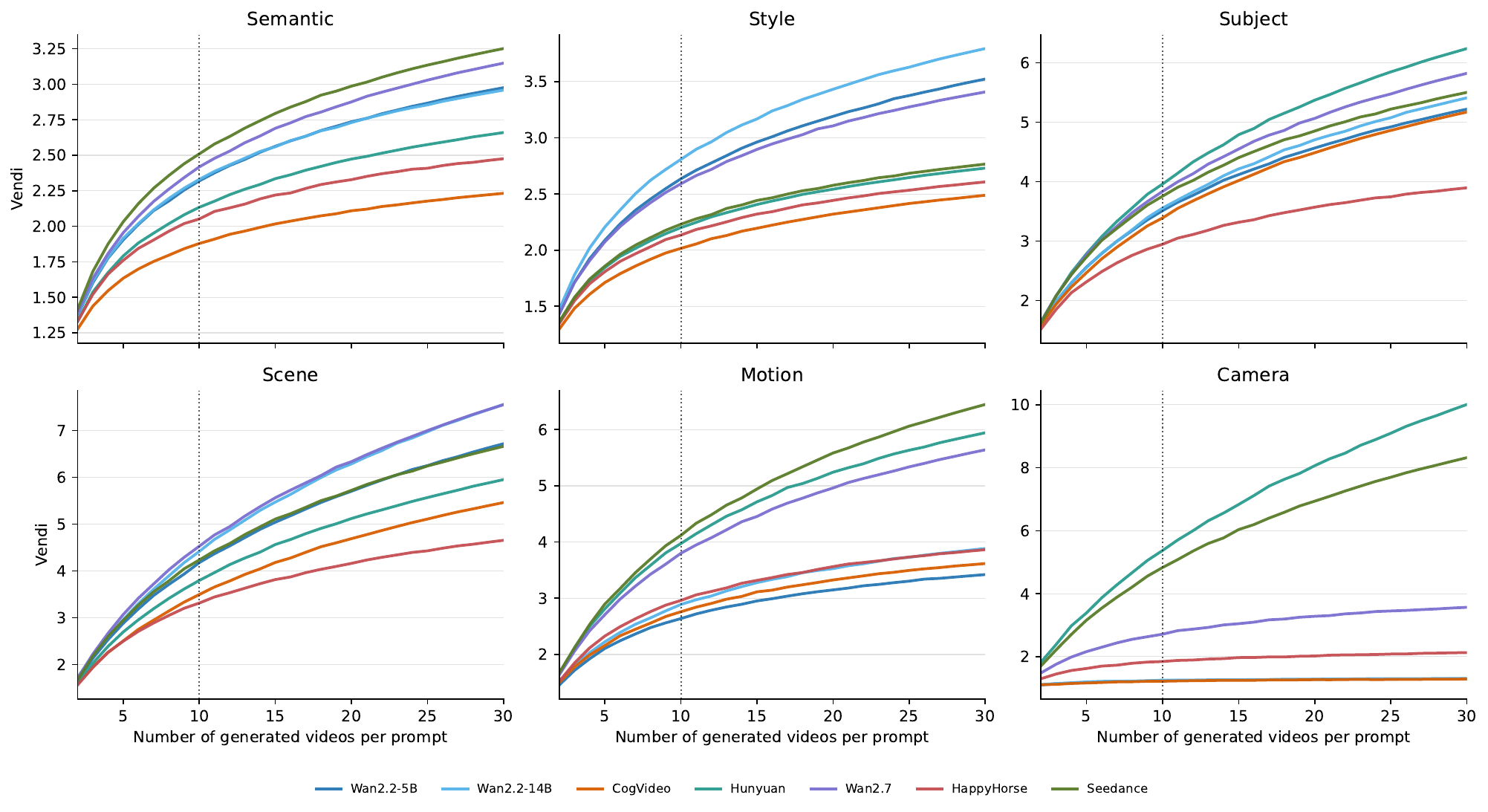}
\end{minipage}
\caption{Sensitivity to the number of generated videos. Left: MPD. Right: raw Vendi Score.}
\label{fig:app4-generation-count}
\end{figure*}

\begin{figure*}[h]
    \centering
    \includegraphics[width=\textwidth]{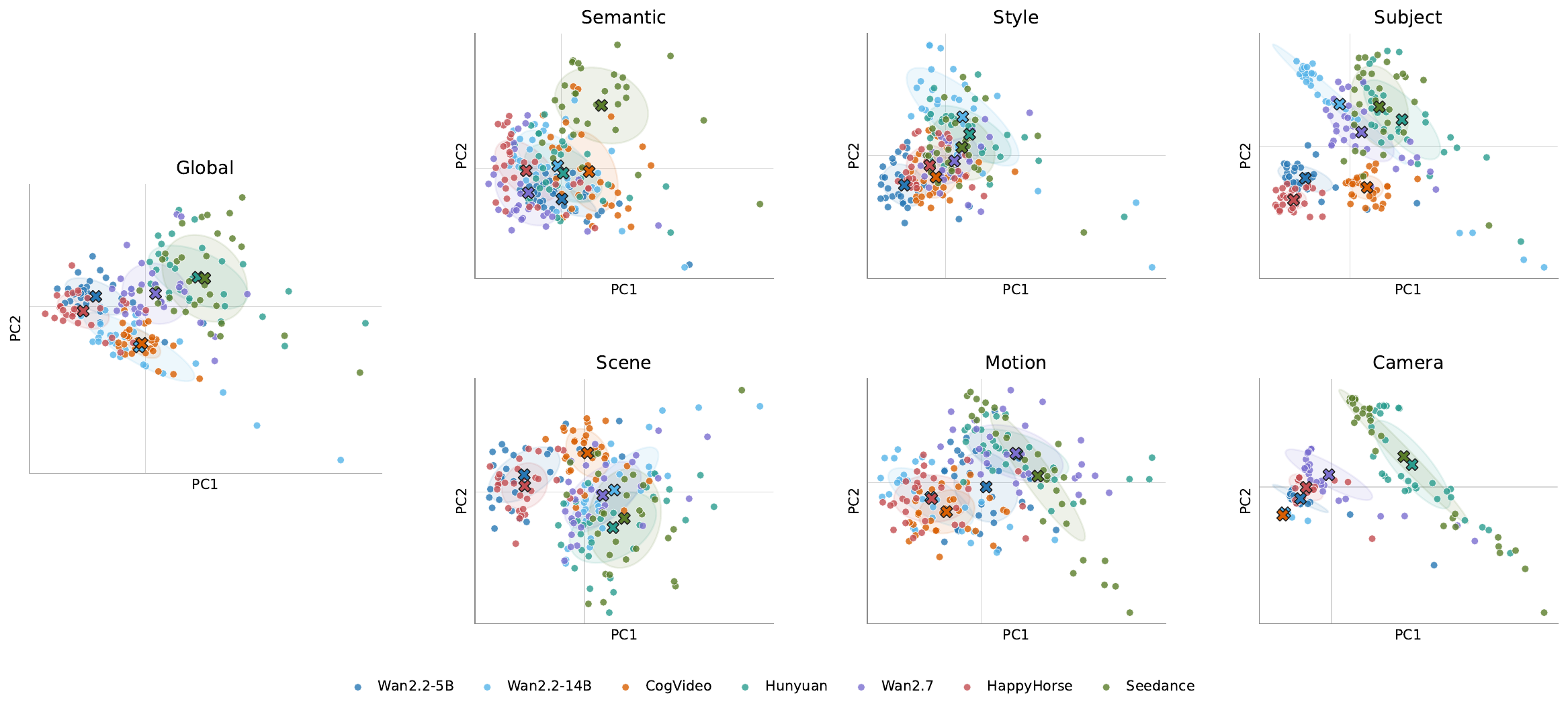}
    \caption{Each point represents a generated video. PCA visualization illustrates both intra-model and inter-model homogenization across six diversity dimensions and the global profile.}
    \label{fig:pca_num}
\end{figure*}

To verify that generating 10 videos per prompt is sufficient, we randomly select 10 prompts and generate 30 videos for each.
In Figure \ref{fig:app4-generation-count}, the models retain the same six-dimensional ordering from $N=10$ to $N=30$ on MPD and Vendi. The MPD scores also remain stable.
Since Vendi Score is equivalent to the effective number of distinct samples, the raw score increases with set size but at a diminishing rate, while model ordering remains stable. Figure \ref{fig:pca_num} visualizes the intra-model and inter-model homogenization by PCA.

\subsection{Additional Information about DiVid Framework}
\label{app:eval-pipeline-final}

Each group of 10 videos takes approximately 30 seconds for the full six-dimension evaluation on one A100 GPU.
All dimensions follow the same set-level interface: each video is mapped to a dimension-specific representation, the representations form a pairwise kernel, and the kernel is aggregated over the videos generated for the same prompt. Encoders and geometric hyperparameters are fixed for all models.

\subsubsection{Details on Diversity Metrics}

\paragraph{TCE and TIE references.}
For comparison with prior global diversity metrics, we follow the video protocol of TCE/TIE \citep{kazimi2026dppgrpo}. Since truncated covariance entropy is sensitive to the number of samples and retained eigenvalues, our TCE/TIE values, computed from ten videos per prompt, are used only for within-benchmark comparison and are not numerically comparable to prior results based on twenty-video sets.

\paragraph{Semantic.}
We uniformly sample eight frames from each video and encode them with CLIP ViT-L/14. After frame averaging and $\ell_2$ normalization, the video embeddings form $X$, while the normalized prompt embedding is $z$. We remove the prompt-alignment direction using
\begin{equation}
K_{\mathrm{sem}} = XX^\top - (Xz)(Xz)^\top .
\end{equation}
Negative eigenvalues caused by numerical error are clipped, and the kernel is normalized by its diagonal. The pairwise distance is $1-K_{ij}$. This operation reduces shared prompt alignment but does not remove all semantic dependence on the prompt.

\paragraph{Style diversity.}
We uniformly sample eight frames per video and extract 2048-dimensional InceptionV3 pool3 features. Frame features are individually $\ell_2$-normalized, averaged over time, and normalized again to form a video-level descriptor. Pairwise cosine similarities define the style kernel, whose mean pairwise distance measures diversity in global appearance, including composition, texture, color, and lighting.

\paragraph{Subject diversity.}
GroundingDINO searches up to 32 frames for an initialization subject box, and SAM2 propagates the mask forward and backward. A track requires at least three valid mask frames. Multi-subject videos are processed per subject and aggregated using mean mask area as the weight.
For each valid frame, we crop the tight subject box and replace pixels outside the mask with the frame-wise RGB mean to reduce background leakage. DINOv2 $[\mathrm{CLS};\text{mean patch}]$ features are averaged over time and normalized before the pairwise kernel is computed.

\paragraph{Scene diversity.}
We suppress the annotated foreground using the union of the tracked subject masks, replacing subject pixels with a Gaussian-blurred version of the frame (radius $12$). Each background-focused frame is represented by the concatenation of the DINOv2 class token and the mean patch token. Frame features are averaged over time and $\ell_2$-normalized to obtain a video descriptor. Pairwise cosine similarities define the scene kernel, and scene diversity is measured by its mean pairwise distance.

\paragraph{Motion diversity.}
CoTracker is initialized at four temporal anchors with a $10\times10$ subject grid and a $12\times12$ background grid. For each adjacent frame pair, a robust global transform $H_t$ is fitted from background tracks, and subject motion is compensated as
\begin{equation}
r_{p,t}=x_{p,t+1}-H_t(x_{p,t}).
\end{equation}
A relative similarity transform fitted to the compensated subject tracks yields translation, scale, and rotation signals. Their temporal statistics, path characteristics, and an eight-bin direction histogram form a 24-dimensional trajectory descriptor. In parallel, smoothed subject boxes are mapped to a $256\times256$ canonical canvas. RAFT flow is computed between canonicalized frames, and the fitted rigid similarity flow is removed. The temporal means and standard deviations of residual magnitude, moving ratio, and an eight-bin direction histogram form a 20-dimensional deformation descriptor. After dimension-wise normalization by fixed scales, the two descriptors define RBF kernels with $\sigma=3.0$, which are combined as
\begin{equation}
K_{\mathrm{motion}}
=0.5K_{\mathrm{traj}}+0.5K_{\mathrm{def}}.
\end{equation}
Only videos with valid trajectory and deformation estimates are included.

\paragraph{Camera diversity.}
Background points outside a five-pixel dilation of the subject mask are used to fit a partial-affine transform.
A homography replaces it when sufficiently supported and substantially more accurate, with its center-local motion converted to the same camera parameters.
Each frame pair yields normalized horizontal translation, vertical translation, log scale, and rotation. Mean, standard deviation, mean absolute magnitude, and the 90th percentile over time produce a 16-dimensional descriptor; frozen feature scales and an RBF kernel with $\sigma=2.0$ are used for all models.

\subsubsection{Subject Annotation}
\begin{figure*}[h]
    \centering
    \includegraphics[width=1.0\linewidth]{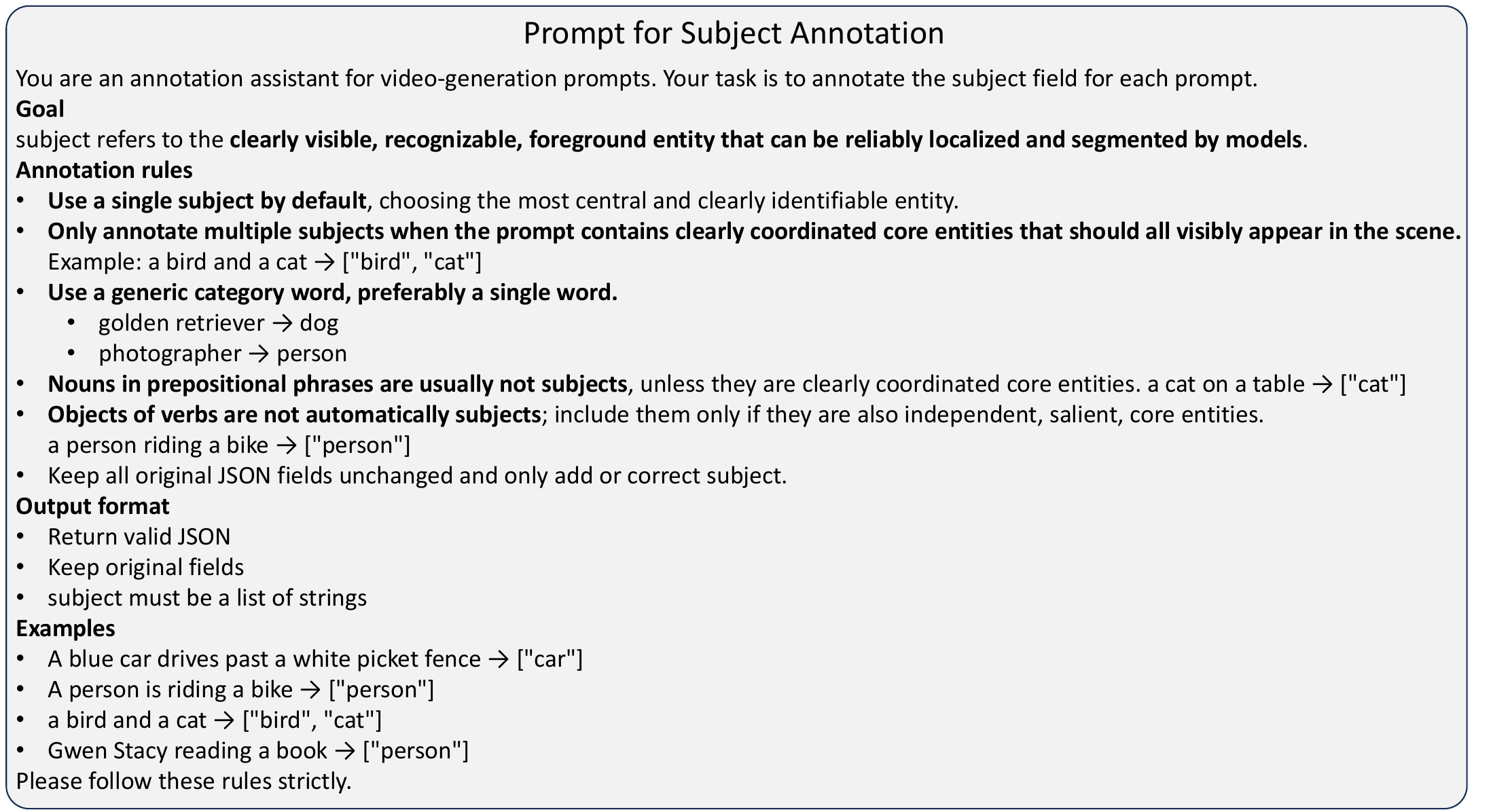}
    \caption{Prompt for Subject Annotation.}
    \label{fig:subject_anno}
\end{figure*}

We define the subject as the foreground entity in a video that is clearly visible, recognizable, and suitable for stable localization and segmentation. By default, we retain only the single most central subject. Multiple subjects are annotated only when the prompt contains clearly coordinated core entities. Subject labels are normalized to generic, single-word category nouns (e.g., \texttt{person}, \texttt{dog}, and \texttt{car}), without modifiers such as color, material, age, or style, and without proper names or character names.
Nouns appearing in prepositional phrases, supporting objects, and background reference objects are usually not annotated as subjects. We use GPT-5 to produce the initial subject annotations for all 206 prompts, and the annotation instruction is shown in Figure~\ref{fig:subject_anno}. For the prompts in Section~\ref{sec:5_explore}, the subject labels are predefined during prompt construction.
After the initial annotation, we asked three annotators to manually review all subject labels. For 94.7\% of the prompts, the GPT-5 annotation was unanimously accepted by all three annotators. For the remaining prompts with disagreement, we adopted a majority-vote strategy and used the label preferred by most annotators as the final subject annotation.

\subsubsection{Set-Level Aggregation}

For a prompt--model set with $N$ valid videos, Mean Pairwise Distance is
\begin{equation}
\mathrm{MPD}(K)=\frac{2}{N(N-1)}\sum_{i<j}(1-K_{ij}).
\end{equation}
It is the expected distance between two uniformly sampled candidates from the observed set. Because each dimension uses a different representation and kernel, raw MPD values are compared only within the same dimension.

Vendi Score normalizes the kernel eigenvalues to sum to one and computes $\exp(-\sum_j\lambda_j\log\lambda_j)$. It provides an effective-mode perspective, whereas MPD is the primary metric because its pairwise interpretation is direct.

\subsection{Validity Checks and Scope}
\label{app:failure}

\paragraph{Automatic validity controls.}
GroundingDINO scans up to 32 sampled frames for an initialization box, using box and text thresholds of 0.30 and 0.25. SAM~2 then propagates each detected subject mask both forward and backward. A video is retained only if every annotated subject has at least three mask frames whose area lies between 0.1\% and 95\% of the image. Multi-subject instances are tracked separately before aggregation. At the prompt level, a factor is excluded rather than assigned zero when fewer than two videos, or fewer than 30\% of the generated videos, retain valid subject masks.
Subject and Scene use complementary leakage controls. For Subject, pixels outside the mask are replaced by the frame-wise RGB mean before the tight crop is encoded. For Scene, the masked foreground is replaced by a Gaussian-blurred region with radius 12. For temporal metrics, Camera samples background points outside a five-pixel dilation of the subject mask, while Motion samples subject points from an eroded interior region. This exclusion margin reduces boundary contamination in both directions.

\paragraph{Robust temporal estimation.}
CoTracker is initialized at four temporal anchors using a $10\times10$ subject grid and a $12\times12$ background grid. Geometry estimation requires at least 12 subject points and 20 background points. Background transformations are fitted with RANSAC and rejected when confidence is below 0.35; subject trajectories additionally require an inlier ratio of at least 0.50. A Camera or trajectory descriptor is retained only when at least two frame pairs and at least 25\% of the attempted pairs are valid. Motion is accepted only when both the background-compensated trajectory and canonical-deformation branches pass their validity checks.
\paragraph{Statistics.}
Among all videos, all annotated subjects receive an initialization detection in 98.85\% of cases; conditional on detection, SAM~2 execution succeeds in 99.94\%, and 96.60\% of all expected videos pass the complete mask-validity gate. Among attempted temporal estimates, 96.70\% of background-transform pairs, 94.21\% of background-valid subject-trajectory pairs, and 98.48\% of canonical-deformation pairs pass their respective gates.
At the final prompt--dimension level, invalid rates are 0.83\% for Subject and Scene, 0.76\% for Motion, and 0.69\% for Camera; the model-level rate across these four dimensions ranges from 0\% to 2.79\%. Invalid records are treated as missing and paired comparisons use only the intersection of valid prompts. Requiring every model to retain at least 8 of 10 valid videos still preserves the leading model in each factor dimension.

\paragraph{Human mask verification.}
To evaluate errors that cannot be detected by internal status flags, we randomly sampled 300 automatically accepted mask overlays across models and prompts. The text prompt and mask overlay were shown to three annotators, with 100 samples assigned to each annotator. A mask passed only when the target subject had no salient omission and the background or other objects had no salient leakage. The aggregated pass rate was 93.3\%; the remaining failures were concentrated in heavy occlusion and multi-object scenes.

\paragraph{Motion sensitivity.}
A controlled 15-prompt set varies only the motion description of the same animal subject. For example, subtle-motion prompts include ''a bear sleeping with its body rising and falling gently'' and ''a bear sleeping while its ears twitch occasionally''; direction variants include ''a bear walking to the left,'' ''walking to the right,'' and ''walking toward the camera''; larger action changes range from ''a bear sleeping peacefully'' to ''standing up and stretching'' and ''running while chasing a salmon.''
In Table \ref{tab:app3-motion-pilot}, motion MPD is lowest for subtle breathing and local twitches, increases for direction or speed changes, and is highest when the action type and amplitude change together.

\begin{table}[t]
\centering
\small
\begin{tabular}{clc}
\toprule
Group & Controlled variation & Motion MPD \\
\midrule
1 & Breathing / ear twitch / deep breath & 0.254 \\
2 & Walk left / right / toward camera & 0.469 \\
3 & Slow / normal / fast locomotion & 0.483 \\
4 & Sleep / stand and stretch / chase & 0.616 \\
5 & Shake / scratch / dig & 0.456 \\
\bottomrule
\end{tabular}
\caption{Motion sensitivity pilot. Each row aggregates three prompts that share a subject and differ in the indicated motion property.}
\label{tab:app3-motion-pilot}

\end{table}

\paragraph{Metric scope.}
Subject, Scene, Motion, and Camera require an explicit, localizable subject. Most prompts in our benchmark specify a focal entity, covering the vast majority of the evaluated prompts and common text-to-video use cases. Prompts outside this scope are excluded by the explicit validity checks.
Motion and Camera are 2D image-plane measurements. The partial-affine Camera descriptor captures dominant global shot motion rather than full 3D camera pose. Strong parallax, dynamic backgrounds, or shot cuts may require richer geometric models, while Motion remains sensitive to tracking quality and does not assess physical correctness. We therefore use these dimensions to diagnose visible motion and camera diversity at the model level.

\paragraph{Representation.}
As with other feature-based diversity metrics, DiVid measures variation through specific encoders. We use representations matched to each dimension and validate them meticulously.
MPD values are therefore interpreted and compared within each dimension rather than across dimensions.
The framework is modular: alternative encoders can be substituted while retaining the same factor definitions, pairwise aggregation, and evaluation protocol.

\subsection{Human Evaluation}
\label{app:human-evaluation}

\paragraph{Evaluation Protocol.}
We recruited six trained annotators and randomly sampled 50 prompts from different categories. For each prompt, three disjoint model pairs were randomly drawn from all seven models, producing 150 unique prompt--model-pair items. Each pair was judged on six diversity dimensions using a five-level ordered scale: \emph{A clearly more diverse} / \emph{A slightly more diverse} / \emph{about the same} / \emph{B slightly more diverse} / \emph{B clearly more diverse}. Each annotator evaluated 40 model pairs, resulting in 240 valid judgments that ensure full coverage of all 150 items, with 90 items evaluated by two annotators to allow for inter-annotator agreement analysis. Model identity, file names, left/right position, and within-set video order were hidden or randomized throughout.
Annotators also rated on factor-level faithfulness across four dimensions (Subject, Scene, Motion, and Camera). Specifically, given the prompt's description for each factor, annotators judged whether the generated videos faithfully followed the corresponding constraint on a 1--5 scale; N/A was selected when the prompt did not specify that factor. Strict requires the automatic and human scores to be identical, while Direction counts agreement when both scores are positive (4--5), both are negative (1--2), or both are N/A.

\paragraph{Quality Control.}
All annotators hold at least a bachelor's degree and completed a unified training session covering the judgment criteria for each dimension before the formal evaluation began. We explicitly train annotators to focus solely on diversity variations in the specific dimension, avoiding confusion.
For motion-related dimensions (Motion and Camera), the annotation interface enforced a minimum playback requirement: annotators must have watched at least 80\% of the total video duration before submission was enabled.
Appearance dimensions (Subject, Scene) and motion dimensions (Motion, Camera) were presented on separate screens to minimize cross-dimension interference.
To monitor annotation reliability, we inserted 2--3 gold standard items per batch of 40 judgments. Each gold item consisted of one set containing 2--3 identical duplicate videos, and the other contained 10 manually curated, clearly distinct videos spanning diverse content. Across all annotators and gold items, every response correctly identified the set of distinct videos as clearly more diverse. These checks confirm that annotators remained attentive and applied the diversity criteria consistently throughout the study.

\begin{table}[h]
\centering
\small
\begin{tabular}{lrrr}
\toprule
Dimension & Spearman $\rho$ & Kendall $\tau_b$ & Kripp. $\alpha$ \\
\midrule
Semantic & 0.604 & 0.539 & 0.703 \\
Style & 0.563 & 0.522 & 0.682 \\
Subject & 0.549 & 0.503 & 0.679 \\
Scene & 0.541 & 0.489 & 0.680 \\
Motion & 0.531 & 0.471 & 0.692 \\
Camera & 0.556 & 0.502 & 0.686 \\
\bottomrule
\end{tabular}
\caption{Human evaluation for 6-dimension diversity. All correlations have $p<0.001$.}
\label{tab:app3-human-diversity}

\end{table}

\begin{table}[h]
\centering
\small
\begin{tabular}{lrrrr}
\toprule
Factor & Strict & Direction  & Kripp. $\alpha$ \\
\midrule
Subject & 63.4\% & 85.4\%  & 0.699 \\
Scene & 65.4\% & 88.9\%  & 0.705 \\
Motion & 63.7\% & 90.1\% & 0.742 \\
Camera & 69.3\% & 90.3\%  & 0.766 \\
\bottomrule
\end{tabular}
\caption{Human evaluation for factor-specific faithfulness.}
\label{tab:app3-human-faithfulness}

\end{table}

\paragraph{Results.}
In Table \ref{tab:app3-human-diversity} and \ref{tab:app3-human-faithfulness}, all six diversity metrics correlate positively with human pairwise preferences, with Spearman $\rho=0.531$--$0.604$ and Kendall $\tau_b=0.471$--$0.539$ ($p<0.001$). Factor-faithfulness direction agreement is 85.4\%--90.3\%.
Krippendorff's $\alpha$ exceeds the commonly accepted reliability threshold of 0.667.
The results support model-level ranking and diagnostic use.

\subsection{Utility of Factor Level Metrics}
\label{app:diagnostic-utility}

We test whether each factor retains information that the other three cannot recover without claiming that the factors are statistically independent or form a unique taxonomy.
The analysis uses the 195 prompts valid for all four factors and all seven models (1,365 model--prompt records). Within each prompt, MPD values are converted to seven-model rank percentiles. Held-out prediction uses prompt-grouped 10-fold cross-validation, so records from one prompt never appear in both train and test folds.

\begin{table*}[h]
\centering
\small
\begin{tabular}{lrrrr}
\toprule
Factor & VIF & Grouped-CV $R^2$ (95\% CI) & Unexplained & Top-model change (95\% CI) \\
\midrule
Subject & 1.73 & 0.419 [0.363, 0.471] & 58.1\% & 25.6\% [19.5, 31.8] \\
Scene & 1.73 & 0.418 [0.363, 0.470] & 58.2\% & 19.5\% [13.8, 25.1] \\
Motion & 1.39 & 0.274 [0.222, 0.326] & 72.6\% & 22.6\% [16.9, 28.7] \\
Camera & 1.37 & 0.266 [0.212, 0.320] & 73.4\% & 26.7\% [20.5, 32.8] \\
\bottomrule
\end{tabular}
\caption{Incremental diagnostic utility. Grouped-CV predicts one factor from the other three. Top-model change compares the prompt-level top set before and after deleting the factor.}
\label{tab:app3-factor-validity}

\end{table*}

In Table \ref{tab:app3-factor-validity}, the other three factors explain only 26.6\%--41.9\% of held-out variance. Removing one factor changes the disjoint top-model set for 19.5\%--26.7\% of prompts. The temporary equal-rank aggregation used for this deletion test is a sensitivity probe, not a benchmark-wide diversity score.

\begin{table}[h]
\centering
\small
\begin{tabular}{lrr}
\toprule
Target factor & Prompts retrieved & Pair count \\
\midrule
Subject & 122/150 (81.3\%) & 286 \\
Scene & 112/150 (74.7\%) & 262 \\
Motion & 138/150 (92.0\%) & 484 \\
Camera & 140/150 (93.3\%) & 521 \\
\bottomrule
\end{tabular}
\caption{Factor-selective pair retrieval on VBench. The target response is in the top 10\%, while all other factor responses are no higher than their medians.}
\label{tab:app3-factor-retrieval}

\end{table}

In Table \ref{tab:app3-factor-retrieval}, under the primary retrieval rule, a target-factor distance must be in the top 10\% while all other factor distances remain below their medians within the same model--prompt group. Every factor retrieves such selective pairs for at least 74.7\%  prompts. This instance-level evidence complements the aggregate prediction tests by showing that each metric can expose changes not simultaneously flagged by the other three.

Together, the results support that the four metrics provide non-interchangeable diagnostic signals in the evaluated benchmark.

\section{Additional Results for the Main Benchmark Evaluation}
\label{app:main-evaluation}

This appendix reports the statistical details, ablations and sensitivity analyses omitted from Section~\ref{sec:main-eval}.

\subsection{Complete Model-Level Statistics}
\label{app:complete-results-final}

\paragraph{Prompt-paired significance analysis.}
To complement the model-level estimates in Table~\ref{tab:app4-diversity-first} and Table~\ref{tab:app4-diversity-second}, we performed prompt-paired inference for the central rank reversal reported in the main text. The prompt was treated as the unit of analysis; each contrast retained prompts with valid scores for both models and used their prompt-wise MPD differences. We estimated 95\% confidence intervals with 20{,}000 paired prompt-bootstrap resamples and applied two-sided Wilcoxon signed-rank tests, followed by Benjamini--Hochberg correction over all $6\times21=126$ dimension--model-pair comparisons.

\begin{table*}[h]
\centering
\scriptsize
\begin{tabular}{lccc}
\toprule
Model & Semantic & Style & Subject \\
\midrule
Wan2.2-5B
& $0.169\pm0.067\ [0.160,0.178]$
& $0.234\pm0.060\ [0.226,0.243]$
& $0.447\pm0.159\ [0.425,0.468]$ \\

Wan2.2-14B
& $0.144\pm0.073\ [0.134,0.154]$
& $0.224\pm0.066\ [0.215,0.233]$
& $0.412\pm0.173\ [0.388,0.436]$ \\

CogVideo
& $0.142\pm0.058\ [0.134,0.150]$
& $0.214\pm0.068\ [0.205,0.223]$
& $0.414\pm0.146\ [0.394,0.434]$ \\

Hunyuan
& $0.142\pm0.060\ [0.134,0.150]$
& $0.197\pm0.063\ [0.188,0.205]$
& $0.392\pm0.163\ [0.370,0.414]$ \\

Wan2.7
& $0.158\pm0.068\ [0.149,0.168]$
& $0.201\pm0.058\ [0.193,0.209]$
& $0.397\pm0.161\ [0.375,0.420]$ \\

HappyHorse
& $0.093\pm0.054\ [0.086,0.101]$
& $0.123\pm0.042\ [0.117,0.129]$
& $0.250\pm0.141\ [0.231,0.269]$ \\

Seedance
& $0.160\pm0.080\ [0.150,0.171]$
& $0.162\pm0.049\ [0.156,0.169]$
& $0.380\pm0.151\ [0.359,0.401]$ \\
\bottomrule
\end{tabular}
\caption{Complete model-level MPD statistics for the Semantic, Style and Subject dimensions. Each cell reports the mean $\pm$ prompt-level standard deviation and the 95\% confidence interval obtained from 20{,}000 prompt-bootstrap resamples.}
\label{tab:app4-diversity-first}
\end{table*}

\begin{table*}[h]
\centering
\scriptsize
\begin{tabular}{lccc}
\toprule
Model & Scene & Motion & Camera \\
\midrule
Wan2.2-5B
& $0.426\pm0.140\ [0.407,0.446]$
& $0.326\pm0.104\ [0.311,0.340]$
& $0.097\pm0.151\ [0.077,0.118]$ \\

Wan2.2-14B
& $0.394\pm0.154\ [0.374,0.416]$
& $0.344\pm0.109\ [0.329,0.359]$
& $0.148\pm0.235\ [0.117,0.181]$ \\

CogVideo
& $0.377\pm0.124\ [0.359,0.394]$
& $0.323\pm0.089\ [0.311,0.336]$
& $0.045\pm0.119\ [0.030,0.063]$ \\

Hunyuan
& $0.378\pm0.135\ [0.360,0.397]$
& $0.433\pm0.093\ [0.420,0.446]$
& $0.490\pm0.222\ [0.458,0.520]$ \\

Wan2.7
& $0.390\pm0.130\ [0.373,0.408]$
& $0.413\pm0.090\ [0.401,0.425]$
& $0.303\pm0.225\ [0.272,0.334]$ \\

HappyHorse
& $0.223\pm0.106\ [0.208,0.237]$
& $0.326\pm0.096\ [0.313,0.339]$
& $0.193\pm0.188\ [0.168,0.219]$ \\

Seedance
& $0.364\pm0.132\ [0.346,0.382]$
& $0.425\pm0.091\ [0.412,0.437]$
& $0.463\pm0.236\ [0.430,0.495]$ \\
\bottomrule
\end{tabular}
\caption{Complete model-level MPD statistics for the Scene, Motion, and Camera dimensions. Each cell reports the mean $\pm$ prompt-level standard deviation and the 95\% confidence interval obtained from 20{,}000 prompt-bootstrap resamples.}
\label{tab:app4-diversity-second}
\end{table*}

\begin{table*}[h]
\centering
\small
\begin{tabular}{llrrrr}
\toprule
Dimension & Positive paired contrast & $N$ & $\Delta$MPD & 95\% bootstrap CI & Global BH $q$ \\
\midrule
Semantic & Wan2.2-5B $-$ Hunyuan & 206 & 0.027 & [0.019, 0.035] & $9.1\times10^{-11}$ \\
Style & Wan2.2-5B $-$ Hunyuan & 206 & 0.038 & [0.030, 0.046] & $1.9\times10^{-16}$ \\
Subject & Wan2.2-5B $-$ Hunyuan & 203 & 0.054 & [0.037, 0.072] & $1.2\times10^{-8}$ \\
Scene & Wan2.2-5B $-$ Hunyuan & 203 & 0.047 & [0.030, 0.063] & $4.6\times10^{-8}$ \\
Motion & Hunyuan $-$ Wan2.2-5B & 204 & 0.107 & [0.089, 0.125] & $5.0\times10^{-20}$ \\
Camera & Hunyuan $-$ Wan2.2-5B & 204 & 0.394 & [0.362, 0.424] & $9.8\times10^{-33}$ \\
\bottomrule
\end{tabular}
\caption{Prompt-paired validation of the central Wan2.2-5B--Hunyuan rank reversal. $\Delta$MPD is positive for the first-named model. Confidence intervals are based on 20{,}000 paired prompt-bootstrap resamples; $q$ values are obtained from two-sided Wilcoxon signed-rank tests with Benjamini--Hochberg correction across all 126 dimension--model-pair comparisons.}
\label{tab:app4-rank-reversal-significance}
\end{table*}

Table~\ref{tab:app4-rank-reversal-significance} confirms the central cross-dimensional reversal. Relative to Hunyuan, Wan2.2-5B achieves higher MPD across all four content dimensions, with paired differences of 0.027--0.054 and confidence intervals that exclude zero ($q<4.7\times10^{-8}$). Conversely, Hunyuan exceeds Wan2.2-5B by 0.107 in Motion and 0.394 in Camera (both $q<5.0\times10^{-20}$). These prompt-paired results statistically substantiate the opposing content-versus-temporal diversity profiles reported in the main text. Together with the observation that six of the seven models span at least three rank positions across the six dimensions, they show that model diversity cannot be represented by a single global ordering and must instead be characterized dimension by dimension.

\subsection{Real-Video Retrieval Diversity Reference}
\label{sec:real_video_reference}

Video generation models are commonly trained on large-scale video--text datasets \citep{bain2021frozen,nan2025openvid}.
To provide a reference for existing diversity metrics and estimate diversity exhibited by real web videos, we design a retrieval-based pipeline over approximately one million captions from OpenVid-1M \citep{nan2025openvid}.
For each evaluation prompt, BGE-M3 \citep{chen2024m3embedding} first retrieves the top 50 captions; we then rerank them by CLIP \citep{radford2021clip} video--text similarity computed over eight uniformly sampled frames, and evaluate the top 10 using the same frame-sampling and DiVid protocols as for generated videos.

\begin{table*}[h]
\centering
\resizebox{\linewidth}{!}{
\begin{tabular}{lccccccccccc}
\toprule
Source & AQ & IQ & TF & MS & Semantic & Style & Subject & Scene & Motion & Camera & Six-dim. Avg. \\
\midrule
Real videos
& 0.574 & 0.707 & 0.980 & 0.991
& 0.239 (1) & 0.261 (1) & 0.523 (1) & 0.540 (1)
& 0.407 (4) & 0.243 (4) & 0.369 (1) \\
\bottomrule
\end{tabular}}
\caption{Quality and diversity of the retrieved real-video reference. Diversity ranks, shown in parentheses, are computed jointly with the seven evaluated generation models.}
\label{tab:real-video-reference}
\end{table*}

As shown in Table~\ref{tab:real-video-reference}, the retrieved videos achieve quality scores comparable to those of the evaluated generators and rank first in Semantic, Style, Subject, Scene, and six-dimension average diversity, but only fourth in Motion and Camera. This real-video reference further confirms that diversity is dimension-specific: even broad content variation does not imply equally rich temporal variation. It also suggests that drawing from large-scale web-video data alone may not provide sufficient supervision for motion and camera diversity, motivating dimension-aware data curation and structured temporal controls rather than global data rebalancing.

\subsection{Quality--Faithfulness--Diversity Misalignment}
\label{app:misalignment-final}

\begin{table}[h]
\centering
\small
\setlength{\tabcolsep}{4pt}
\begin{tabular}{lrrrrrr}
\toprule
Metric & Sem. & Sty. & Subj. & Scene & Mot. & Cam. \\
\midrule
IF & -0.214 & -0.679 & -0.786 & -0.429 & 0.393 & 0.500 \\
AQ & 0.179 & -0.179 & -0.393 & -0.036 & 0.286 & 0.214 \\
IQ & -0.071 & -0.214 & -0.357 & 0.000 & 0.000 & 0.071 \\
\bottomrule
\end{tabular}
\caption{Spearman correlations across seven model means.}
\label{tab:app4-model-correlations}

\end{table}

\begin{table}[h]
\centering
\small
\begin{tabular}{lrrr}
\toprule
Metric & Negative & Positive & Total \\
\midrule
IF & 15 & 0 & 15 \\
AQ & 13 & 0 & 13 \\
IQ & 7 & 14 & 21 \\
\midrule
Total & 35 & 14 & 49 \\
\bottomrule
\end{tabular}
\caption{Prompt-level correlations significant after BH-FDR correction across 126 tests.}
\label{tab:app4-prompt-correlations}

\end{table}

In Table \ref{tab:app4-model-correlations} and \ref{tab:app4-prompt-correlations}, across seven model means, the strongest negative association is IF--Subject ($\rho=-0.786$), but IF correlates positively with Motion and Camera. At the prompt level, 49 of 126 model-by-metric-by-dimension correlations remain significant after FDR correction: 35 are negative and 14 are positive.

\subsection{Faithfulness--Diversity Decision}
\label{app:Faithfulness-Diversity}

\begin{table*}[h]
\centering
\small
\begin{tabular}{lccccc}
\toprule
Model & Overall IF & Subject & Scene & Motion & Camera \\
\midrule
Wan2.2-5B & 0.848 & 0.853 & 0.928 & 0.778 & 0.938 \\
Wan2.2-14B & 0.915 & 0.914 & 0.957 & 0.890 & 0.967 \\
CogVideo & 0.839 & 0.813 & 0.919 & 0.790 & 0.971 \\
Hunyuan & 0.893 & 0.898 & 0.937 & 0.842 & 0.977 \\
Wan2.7 & 0.963 & 0.962 & 0.992 & 0.935 & 0.986 \\
HappyHorse & \textbf{0.986} & \textbf{0.984} & \textbf{0.999} & \textbf{0.968} & \textbf{1.000} \\
Seedance & 0.955 & 0.952 & 0.974 & 0.934 & 0.979 \\
\bottomrule
\end{tabular}
\caption{Overall and factor-specific faithfulness.}
\label{tab:app4-factor-faithfulness}

\end{table*}

\subsubsection{Threshold-Gated Diversity}

For a threshold $\tau$, only videos whose matched faithfulness is at least $\tau$ are eligible. To prevent models with more surviving videos from receiving more pairs, equal-count MPD samples exactly $k=3$ or $k=5$ faithful videos per model--prompt group and retains only prompts feasible for all seven models.

Style, Subject, and Scene are led by Wan2.2-5B throughout $\tau\in[0.50,1.00]$, while Motion and Camera are led by Hunyuan. Semantic switches from Wan2.2-5B to Seedance at $\tau=0.85$, but the margin is only 0.0015--0.0021 and Seedance's bootstrap top-1 frequency is 51.8\%--55.2\%; we treat this as a near tie rather than a robust reversal.

\begin{figure*}[h]
\centering
\includegraphics[width=0.8\textwidth]{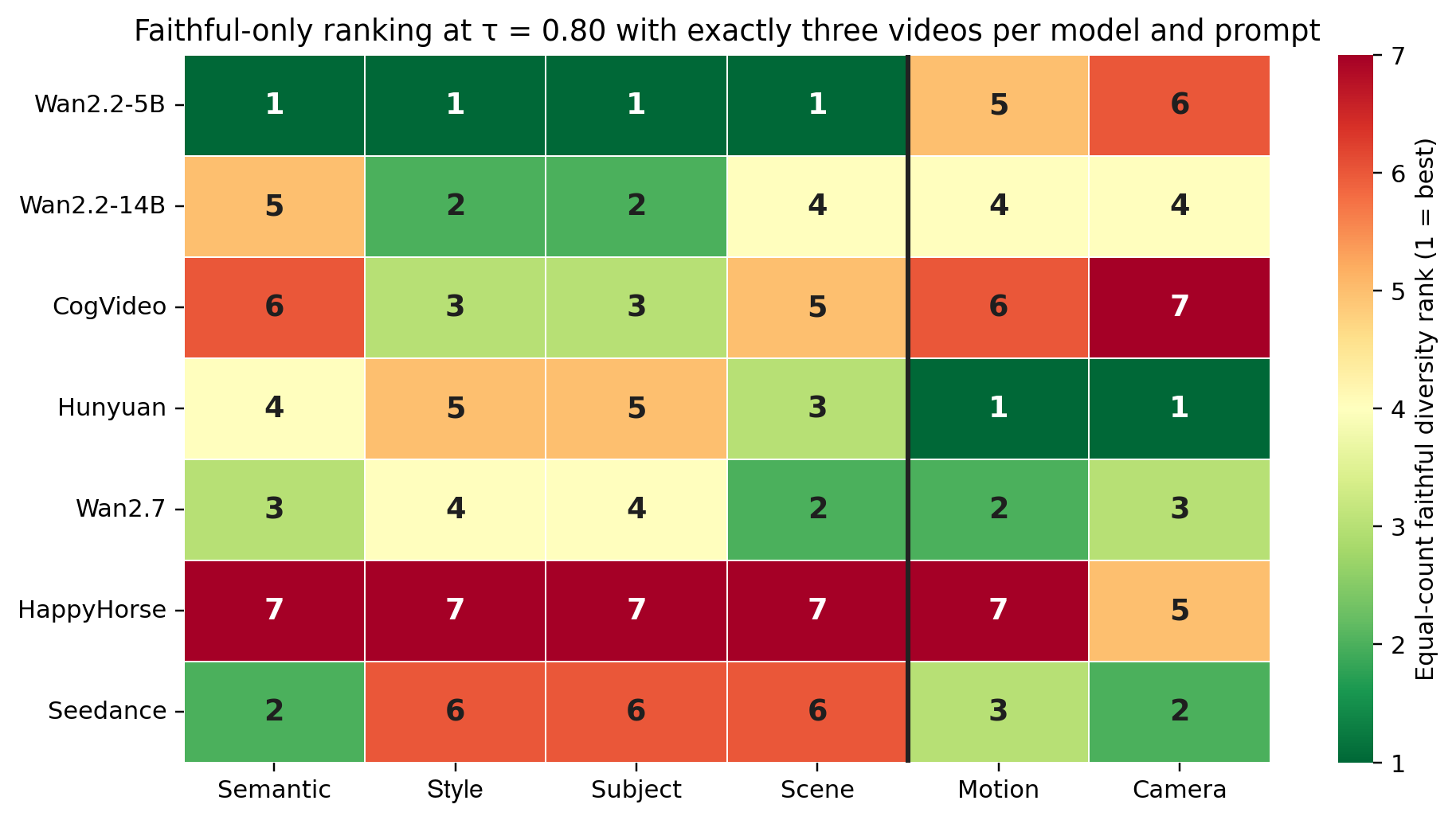}
\caption{Equal-count diversity ranking at the primary setting $\tau=0.8$, $k=3$. Every cell displays the within-dimension model rank.}
\label{fig:app4-equal-count-ranking}
\end{figure*}

\subsubsection{Continuous Preference Weights}

Within each prompt and dimension, matched faithfulness and MPD are independently min--max normalized across the seven models to $\hat F,\hat D\in[0,1]$. If one axis has zero range, it contributes no model distinction. Among Pareto-nondominated models, Balanced Pareto Win Rate selects the model closest to the ideal point using
\begin{equation}
S=1-\sqrt{(1-w_D)(1-\hat F)^2+w_D(1-\hat D)^2}.
\end{equation}
The six-dimension macro curve averages win rates, not raw MPD, and is used only as a recommendation summary.

\begin{figure*}[h]
\centering
\includegraphics[width=0.98\textwidth]{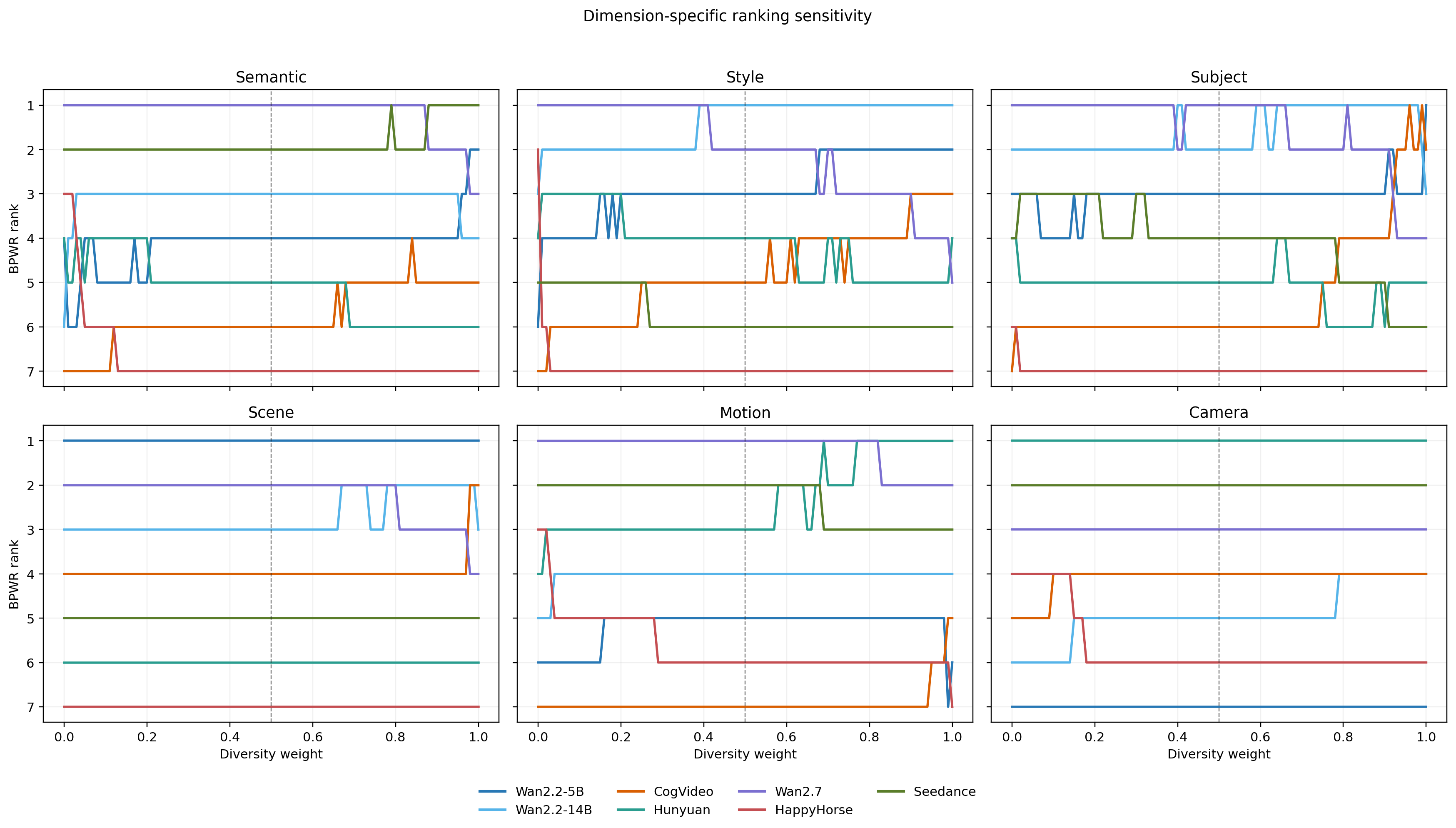}
\caption{Dimension-specific BPWR ranks over the full diversity-weight range.}
\label{fig:app4-bpwr-dimensions}
\end{figure*}

Wan2.7 ranks first for $w_D\in[0,0.83]$, and Hunyuan ranks first for $w_D\in[0.84,1]$; their top-2 coverage widths are 0.98 and 0.94. This recommendation is robust over broad preference ranges but remains dimension dependent, as shown in Figure~\ref{fig:app4-bpwr-dimensions}.

For a separate quality-oriented deployment scenario, we normalize AQ, IQ, and diversity within each prompt and assign weights 0.25/0.25/0.50. This is not the faithfulness--diversity score above: it answers which model balances perceptual quality and diversity when instruction fidelity is not the explicit constraint.

\begin{figure*}[h]
\centering
\includegraphics[width=0.90\textwidth]{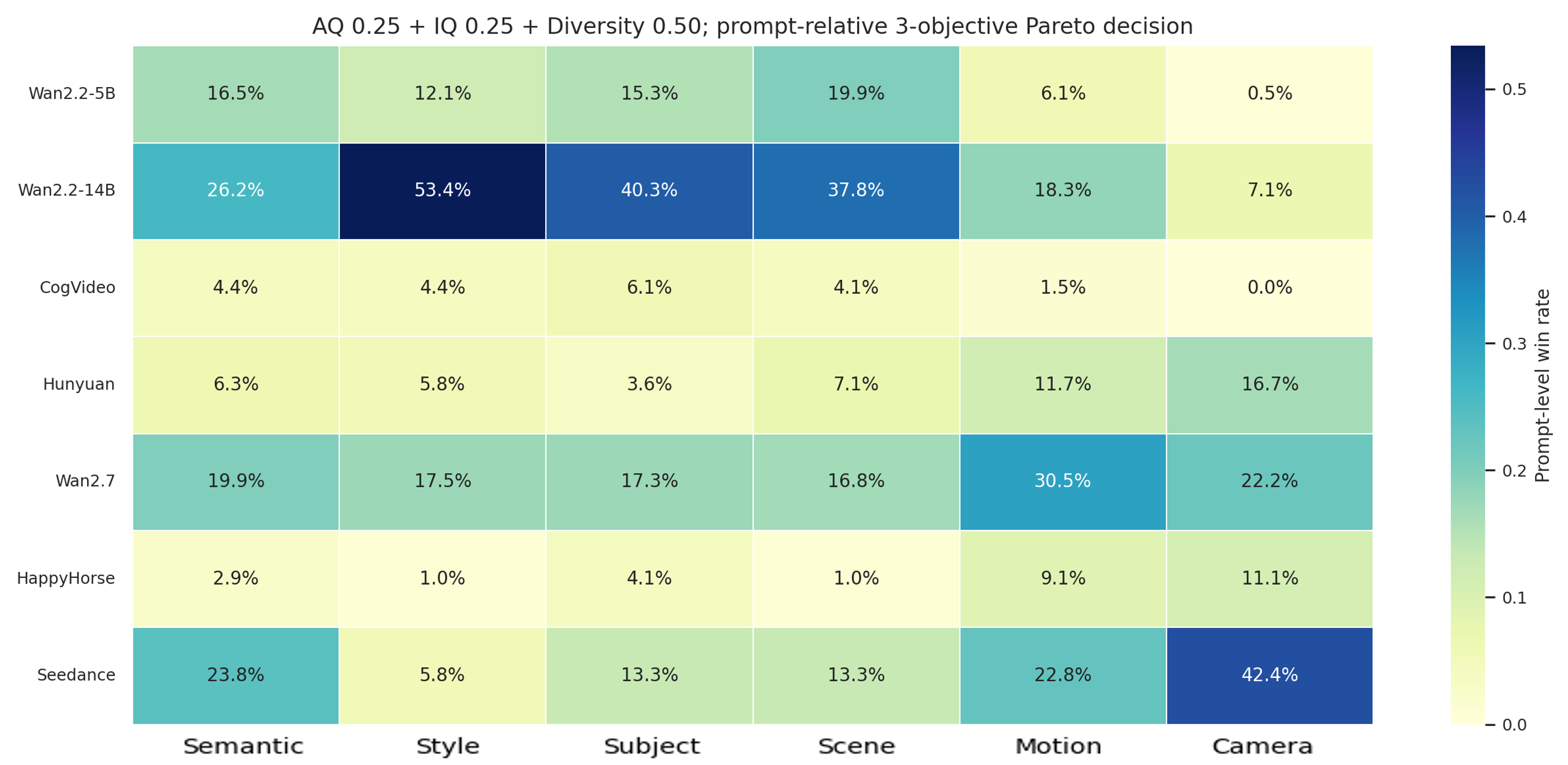}
\caption{Prompt-level quality--diversity decision rates with AQ/IQ/Diversity weights 0.25/0.25/0.50.}
\label{fig:app4-quality-diversity-decision}
\end{figure*}

\subsection{Cross-Prompt-Set Transfer}
\label{app:prompt-transfer}

Tables \ref{tab:app4-vbench-full-results} and \ref{tab:app4-compbench-full-results} show the raw results on VBench and T2V-CompBench. Across the six diversity dimensions, 83.3\% of the 21 model-pair orders are preserved on average. Style and Camera preserve all pairwise orders; Subject, Scene, and Motion show limited top-model changes under compositional prompts.

\begin{table*}[h]
\centering
\resizebox{\textwidth}{!}{%
\begin{tabular}{l rrrrr rr rr rrrr}
\toprule
& \multicolumn{5}{c}{\textbf{Faith / Quality}}
& \multicolumn{2}{c}{\textbf{Prior Metrics}}
& \multicolumn{2}{c}{\textbf{Holistic}}
& \multicolumn{4}{c}{\textbf{Factor Level}} \\
\cmidrule(lr){2-6}
\cmidrule(lr){7-8}
\cmidrule(lr){9-10}
\cmidrule(lr){11-14}
Model & IF & AQ & IQ & TF & MS & TCE & TIE & Semantic & Style & Subject & Scene & Motion & Camera \\
\midrule
Wan2.2-5B
& 0.845 & 0.553 & 0.682 & \textbf{0.982} & 0.990 & \textbf{5.555} & \textbf{13.422} & \textbf{0.184} & \textbf{0.241} & \textbf{0.473} & \textbf{0.450} & 0.317 & 0.083 \\
Wan2.2-14B
& 0.916 & \textbf{0.620} & 0.704 & 0.972 & 0.984 & 4.590 & \underline{13.262} & 0.149 & \underline{0.228} & \underline{0.425} & 0.406 & 0.338 & 0.146 \\
CogVideo
& 0.857 & 0.531 & 0.636 & 0.974 & 0.986 & 4.006 & 12.385 & 0.148 & 0.217 & 0.421 & 0.382 & 0.315 & 0.051 \\
Hunyuan
& 0.899 & 0.549 & 0.670 & 0.965 & \textbf{0.991} & 5.318 & 13.263 & 0.154 & 0.206 & 0.416 & 0.399 & \textbf{0.450} & \textbf{0.504} \\
\midrule
Wan2.7
& \underline{0.965} & 0.572 & \underline{0.731} & 0.972 & 0.985 & 5.177 & 12.765 & 0.170 & 0.206 & 0.414 & \underline{0.408} & 0.422 & 0.308 \\
HappyHorse
& \textbf{0.988} & \underline{0.605} & \textbf{0.736} & \underline{0.978} & \underline{0.991} & 2.806 & 11.487 & 0.099 & 0.129 & 0.272 & 0.238 & 0.337 & 0.202 \\
Seedance
& 0.958 & 0.602 & 0.673 & 0.970 & 0.990 & \underline{5.547} & 13.228 & \underline{0.180} & 0.168 & 0.397 & 0.388 & \underline{0.440} & \underline{0.497} \\
\bottomrule
\end{tabular}%
}
\caption{Complete quality, faithfulness, and diversity profiles on the 150 VBench prompts.}
\label{tab:app4-vbench-full-results}
\end{table*}

\begin{table*}[t]
\centering
\resizebox{\textwidth}{!}{%
\begin{tabular}{l rrrrr rr rr rrrr}
\toprule
& \multicolumn{5}{c}{\textbf{Faith / Quality}}
& \multicolumn{2}{c}{\textbf{Prior Metrics}}
& \multicolumn{2}{c}{\textbf{Holistic}}
& \multicolumn{4}{c}{\textbf{Factor Level}} \\
\cmidrule(lr){2-6}
\cmidrule(lr){7-8}
\cmidrule(lr){9-10}
\cmidrule(lr){11-14}
Model & IF & AQ & IQ & TF & MS & TCE & TIE & Semantic & Style & Subject & Scene & Motion & Camera \\
\midrule
Wan2.2-5B
& 0.858 & 0.518 & 0.645 & \underline{0.974} & 0.986 & \textbf{4.564} & 13.198 & \textbf{0.131} & \textbf{0.216} & \underline{0.376} & \underline{0.363} & 0.349 & 0.134 \\
Wan2.2-14B
& 0.912 & \textbf{0.598} & 0.683 & 0.970 & 0.981 & 4.388 & \underline{13.368} & \underline{0.131} & \underline{0.211} & 0.376 & 0.361 & \underline{0.358} & 0.155 \\
CogVideo
& 0.789 & 0.475 & 0.585 & 0.970 & 0.983 & \underline{4.418} & \textbf{13.407} & 0.127 & 0.207 & \textbf{0.395} & \textbf{0.363} & 0.345 & 0.030 \\
Hunyuan
& 0.876 & 0.517 & 0.641 & 0.965 & \underline{0.991} & 4.351 & 12.546 & 0.109 & 0.173 & 0.329 & 0.323 & 0.389 & \textbf{0.453} \\
\midrule
Wan2.7
& \underline{0.958} & 0.539 & \underline{0.700} & 0.972 & 0.985 & 4.310 & 12.664 & 0.129 & 0.187 & 0.353 & 0.343 & \textbf{0.390} & 0.289 \\
HappyHorse
& \textbf{0.979} & 0.566 & \textbf{0.710} & \textbf{0.977} & \textbf{0.991} & 2.679 & 11.342 & 0.078 & 0.107 & 0.190 & 0.181 & 0.298 & 0.169 \\
Seedance
& 0.947 & \underline{0.577} & 0.652 & 0.969 & 0.989 & 4.140 & 12.724 & 0.109 & 0.147 & 0.334 & 0.298 & 0.382 & \underline{0.369} \\
\bottomrule
\end{tabular}%
}
\caption{Complete quality, faithfulness, and diversity profiles on the 56 T2V-CompBench prompts.}
\label{tab:app4-compbench-full-results}
\end{table*}

\subsection{Cross-Model Complementarity}
\label{app:model-complementarity}

\begin{table}[t]
\centering
\small
\begin{tabular}{llr}
\toprule
Dimension & Most complementary pair & $\Delta_{\mathrm{comp}}$ \\
\midrule
Semantic & CogVideo + HappyHorse & 0.140 \\
Style & CogVideo + HappyHorse & 0.117 \\
Subject & CogVideo + HappyHorse & 0.230 \\
Scene & CogVideo + HappyHorse & 0.257 \\
Motion & Wan2.2-5B + Seedance & 0.139 \\
Camera & CogVideo + Seedance & 0.152 \\
\bottomrule
\end{tabular}
\caption{Model pair with the largest mean complementarity in each dimension.}
\label{tab:app4-complementarity}

\end{table}

For each prompt and model pair, $D_{\mathrm{within}}$ averages the two matched within-model blocks, and $D_{\mathrm{between}}$ averages all cross-model pairs. We define descriptive complementarity as
\begin{equation}
\Delta_{\mathrm{comp}}=D_{\mathrm{between}}-D_{\mathrm{within}}.
\end{equation}
The union improves over the matched within baseline on 97.0\% of prompt--dimension comparisons on average.

\begin{figure*}[t]
\centering
\includegraphics[width=0.98\textwidth]{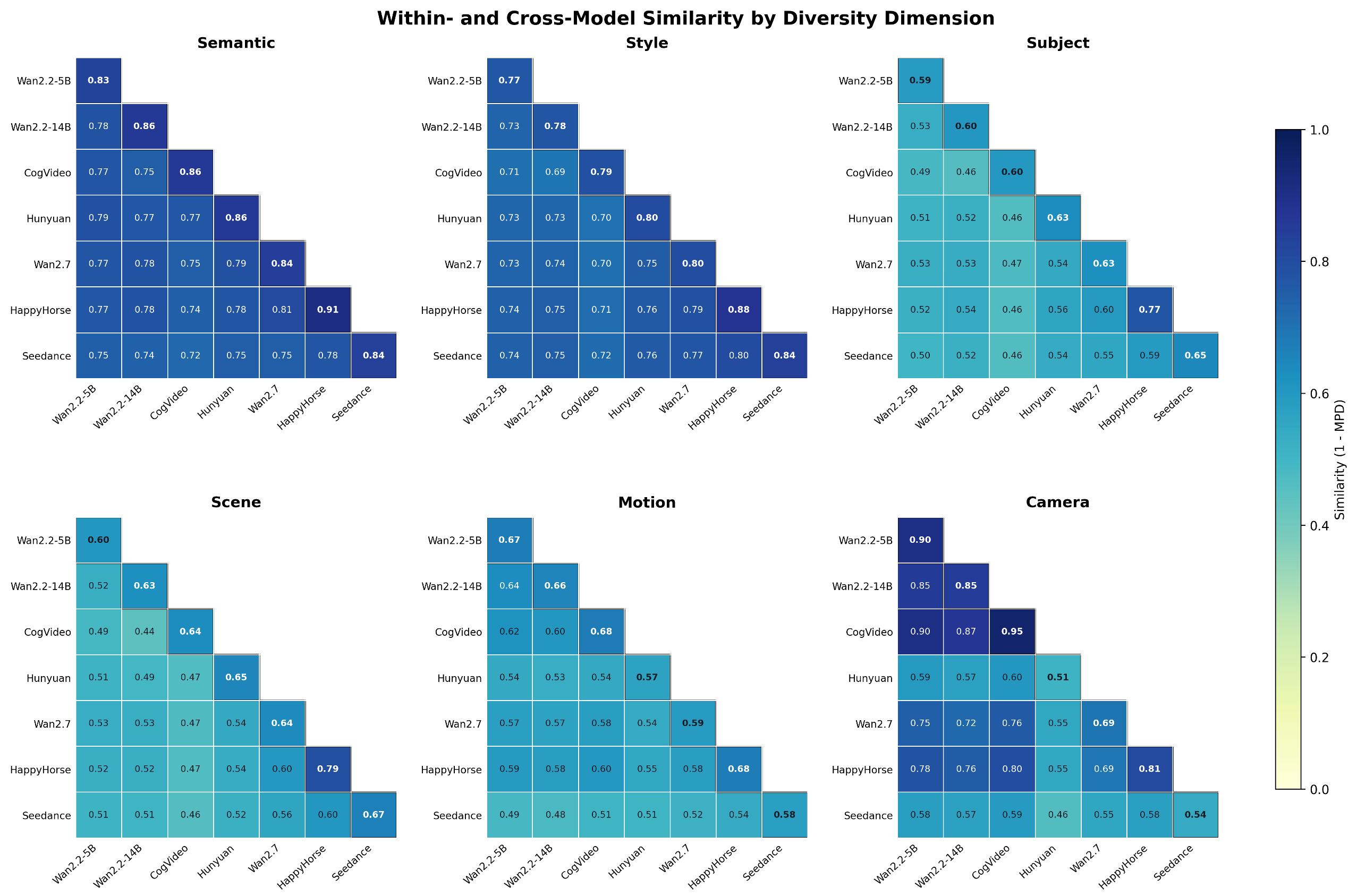}
\caption{intra- and inter-model similarity matrices.}
\label{fig:app4-cross-model-matrices}
\end{figure*}

CogVideo and HappyHorse have relatively low standalone content diversity yet form the most complementary pair in four dimensions. Thus selecting two individually diverse models is not sufficient: routing should also consider whether their default output preferences differ.

\subsection{Dimension-Specific Case Studies}
\label{app:failure-cases}

Global failures concern repeated concepts or appearance; composition failures concern subject or scene collapse; temporal failures concern repeated action or camera policies. Each shows four models selected to expose a clear MPD contrast.

\begin{figure*}[p]
\centering
\textbf{Semantic prompt: tower}\\[2pt]
\includegraphics[width=\textwidth]{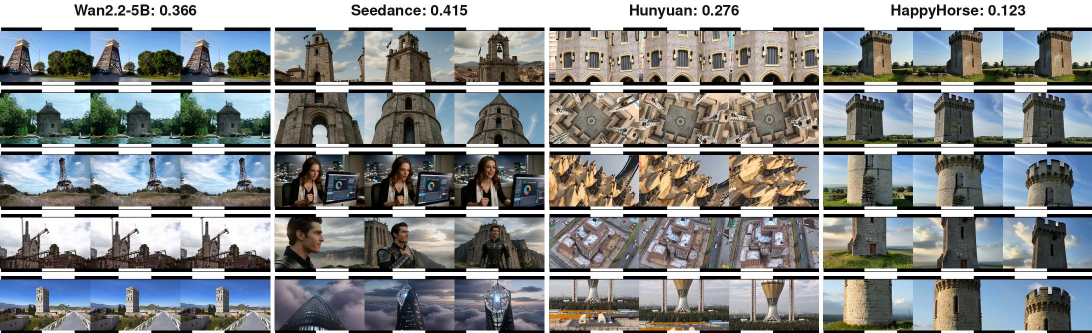}\\[8pt]
\textbf{Style prompt: Two cars collide at an intersection}\\[2pt]
\includegraphics[width=\textwidth]{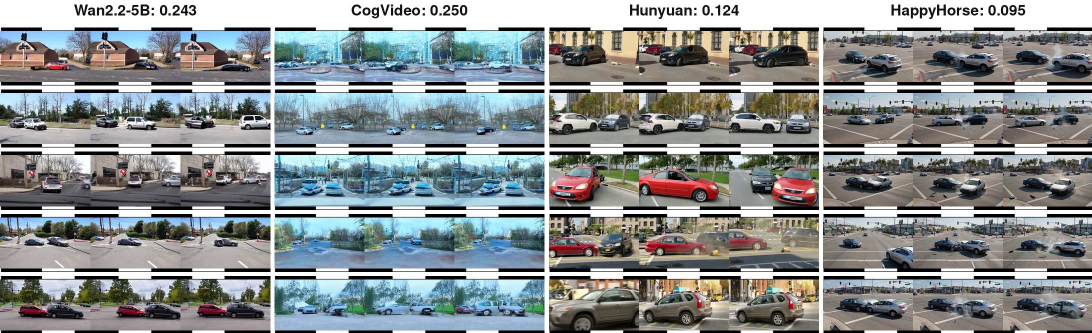}
\caption{Holistic cases.}
\label{fig:app4-cases-global}
\end{figure*}

\paragraph{Holistic.}
Semantic collapse appears when repeated samples preserve nearly the same concept realization, whereas Style collapse preserves color, viewpoint, texture, or layout even when the nominal subject changes. Figure~\ref{fig:app4-cases-global} shows that the benchmark separates these two forms of global homogenization.

\begin{figure*}[p]
\centering
\textbf{Subject prompt: A deer grazes in the meadow as a rabbit hops by}\\[2pt]
\includegraphics[width=\textwidth]{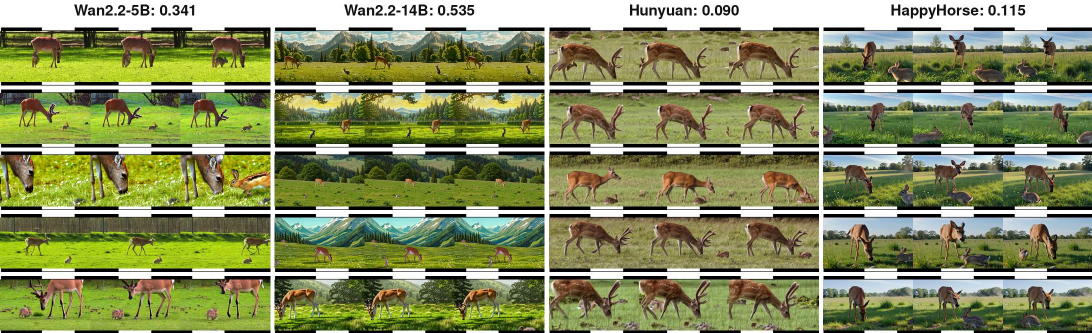}\\[8pt]
\textbf{Scene prompt: A goldfish bowl set on a suitcase}\\[2pt]
\includegraphics[width=\textwidth]{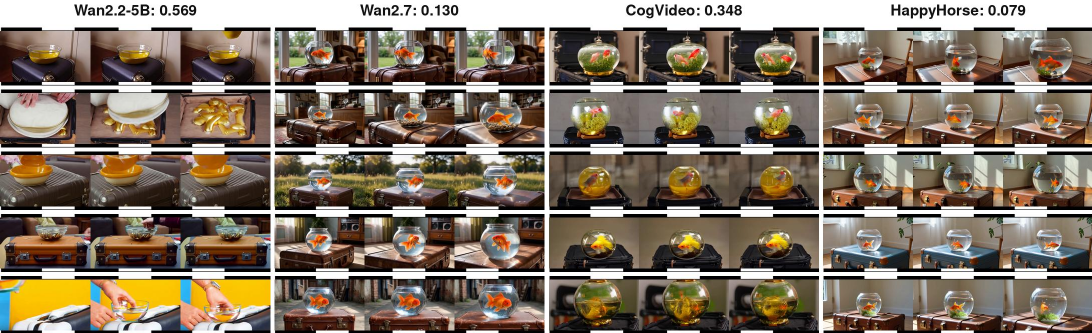}
\caption{Composition cases.}
\label{fig:app4-cases-composition}
\end{figure*}

\paragraph{Composition exploration.}
Subject collapse retains nearly identical foreground identity or appearance, while Scene collapse retains the same environment despite repeated sampling. Figure~\ref{fig:app4-cases-composition} shows the cases.

\begin{figure*}[p]
\centering
\textbf{Motion prompt: A person swimming in ocean}\\[2pt]
\includegraphics[width=\textwidth]{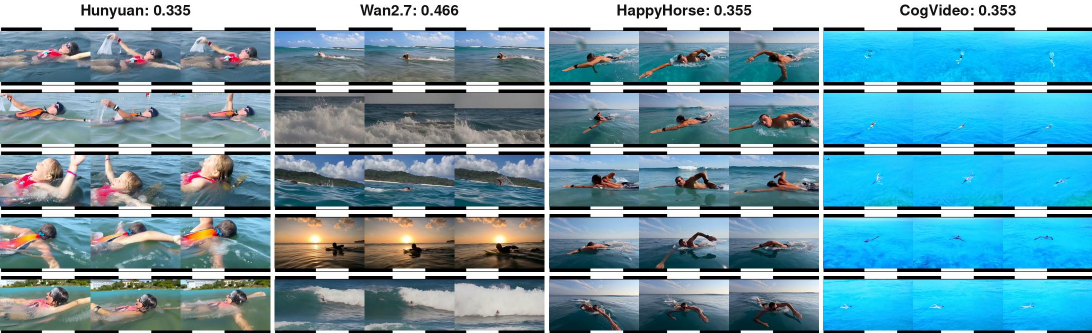}\\[8pt]
\textbf{Camera prompt: family enjoying snack time while sitting in the living room}\\[2pt]
\includegraphics[width=\textwidth]{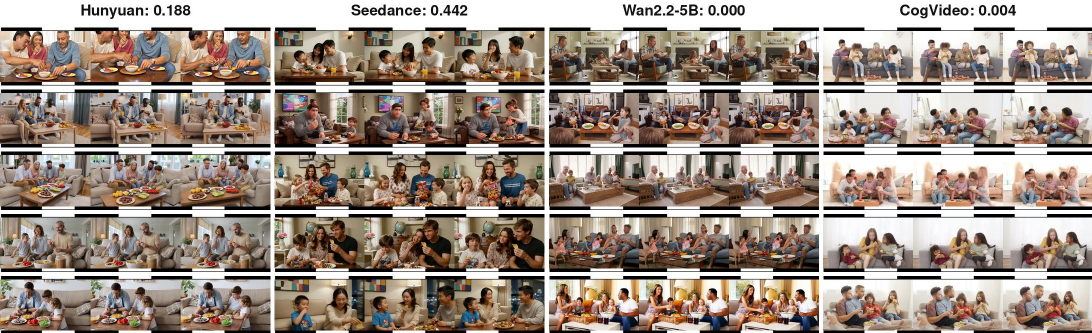}
\caption{Temporal cases.}
\label{fig:app4-cases-temporal}
\end{figure*}

\paragraph{Temporal exploration.}
Motion collapse repeats subject trajectories or deformation, while Camera collapse repeats the shot policy. Figure \ref{fig:app4-cases-temporal} shows the cases.

\section{Supplementary Materials for Section 5}
\label{app:dimension-exploration}

This appendix supplements Section \ref{sec:5_explore} with the controlled-prompt construction, statistical protocol, faithfulness diagnostics, and representative failure modes.

\subsection{Complete Experimental Design and Prompt Construction}
\label{app5:prompt-design}

We use the fully specified prompt as the paired baseline and change only one target dimension at a time. The four settings are defined in Table~\ref{tab:app5-prompt-settings}. \textbf{Specified} provides the reference constraint, \textbf{Omitted} removes the target description, \textbf{Hinted} replaces it with a generic cue, and \textbf{Enumerated} provides five explicit candidates. Subject omission is undefined because a valid generation prompt must contain a subject. Each Enumerated candidate is a different prompt and is generated once.

\begin{table}[t]
\centering
\small
\begin{tabularx}{\columnwidth}{@{}l X X@{}}
\toprule
Setting & Construction & Diagnostic role \\
\midrule
Specified & All four dimensions are explicitly specified. & Paired reference condition. \\
Omitted & Remove the target-dimension description. & Tests autonomous exploration. \\
Hinted & Replace the specific description with a generic cue. & Tests dimension activation. \\
Enumerated & Enumerate target candidates; generate one video per prompt. & Tests candidate realization. \\
\bottomrule
\end{tabularx}
\caption{Controlled prompt settings. Subject omission is undefined because a valid prompt requires a subject.}
\label{tab:app5-prompt-settings}

\end{table}

Specified prompts were first drafted with a common template, then screened by three independent reviewers. A prompt was retained only when the four dimensions admitted multiple plausible alternatives, the target span could be replaced without changing the other spans, and the resulting sentence was grammatical and physically plausible. The controlled set contains 20 prompts covering people, animals, vehicles, indoor and outdoor environments, and urban and natural scenes.

\begin{figure*}[h]
    \centering
    \includegraphics[width=1.0\linewidth]{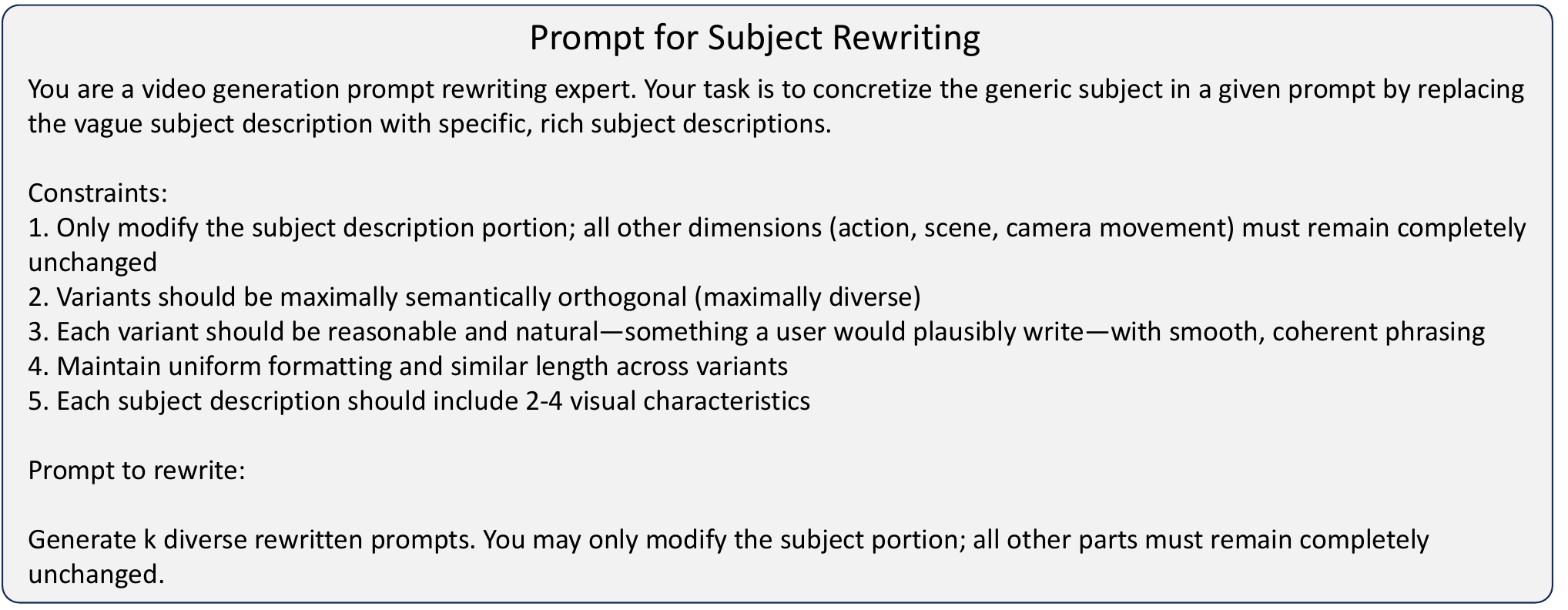}
    \caption{Prompt for Subject Rewriting.}
    \label{fig:rewrite_subject}
\end{figure*}

\begin{figure*}[h]
    \centering
    \includegraphics[width=1.0\linewidth]{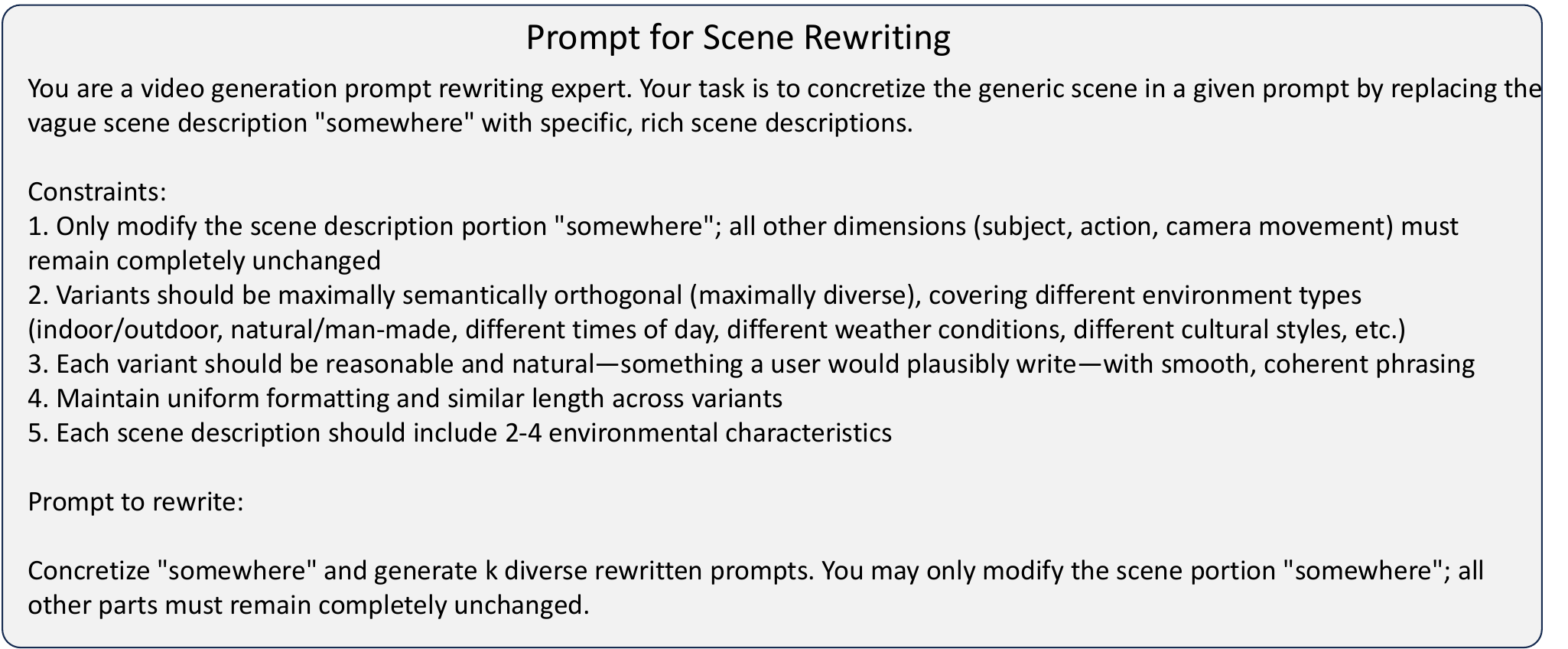}
    \caption{Prompt for Scene Rewriting.}
    \label{fig:rewrite_scene}
\end{figure*}

\begin{figure*}[h]
    \centering
    \includegraphics[width=1.0\linewidth]{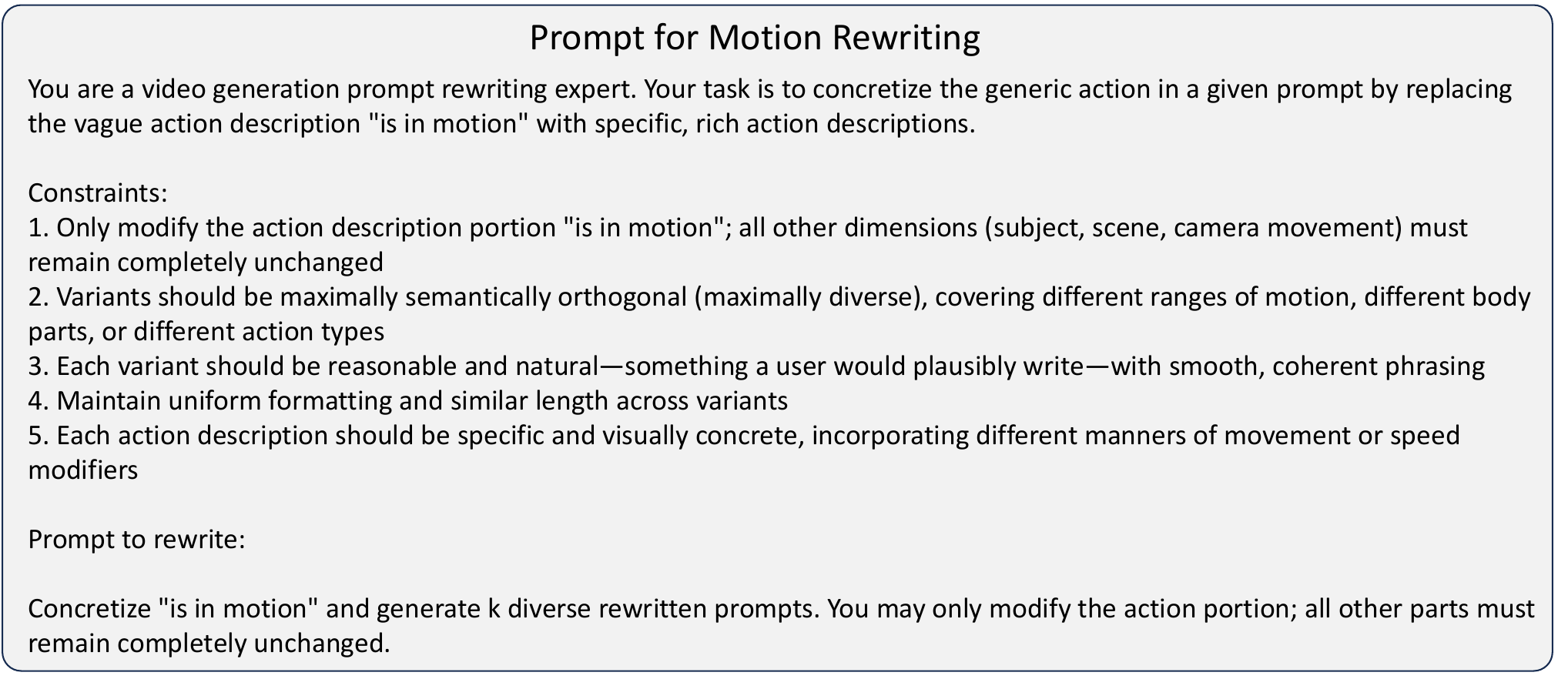}
    \caption{Prompt for Motion Rewriting.}
    \label{fig:rewrite_motion}
\end{figure*}

\begin{figure*}[h]
    \centering
    \includegraphics[width=1.0\linewidth]{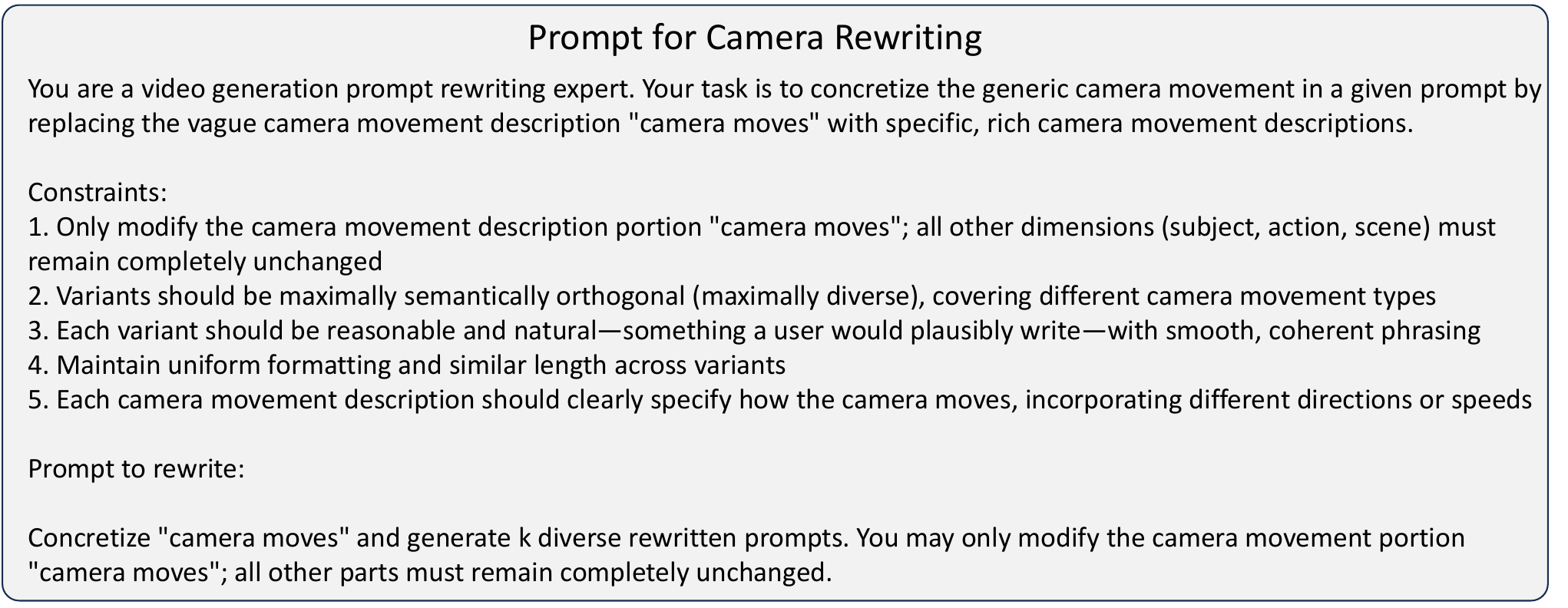}
    \caption{Prompt for Camera Rewriting.}
    \label{fig:rewrite_camera}
\end{figure*}

For example, the golden-retriever prompt is constructed by replacing only one span at a time: the subject becomes \texttt{an animal}, the scene becomes \texttt{somewhere}, the motion becomes \texttt{in motion}, or the camera becomes \texttt{camera moves}. The complete example, including all five Enumerated prompts for each dimension, is given in Table~\ref{tab:app5-prompt-case}.
We use GPT-5 to generate Enumerated prompt. The complete prompts are in Figure \ref{fig:rewrite_subject}, \ref{fig:rewrite_scene}, \ref{fig:rewrite_motion}, \ref{fig:rewrite_camera}.

\begin{table*}[t]
\centering
\scriptsize
\setlength{\tabcolsep}{3pt}

\begin{tabular}{@{}lll p{0.73\textwidth}@{}}
\toprule
Setting & Target & No. & Prompt \\
\midrule
Specified & All & -- & A large fluffy adult golden retriever is resting on a quiet sandy beach at golden hour, static camera \\
Hinted & Subject & -- & An animal is resting on a quiet sandy beach at golden hour, static camera \\
Hinted & Scene & -- & A large fluffy adult golden retriever is resting somewhere, static camera \\
Hinted & Motion & -- & A large fluffy adult golden retriever is in motion on a quiet sandy beach at golden hour, static camera \\
Hinted & Camera & -- & A large fluffy adult golden retriever is resting on a quiet sandy beach at golden hour, camera moves \\
Omitted & Scene & -- & A large fluffy adult golden retriever is resting, static camera \\
Omitted & Motion & -- & A large fluffy adult golden retriever is on a quiet sandy beach at golden hour, static camera \\
Omitted & Camera & -- & A large fluffy adult golden retriever is resting on a quiet sandy beach at golden hour \\
\midrule
Enumerated & Subject & 1 & A large fluffy adult golden retriever is resting on a quiet sandy beach at golden hour, static camera \\
Enumerated & Subject & 2 & A small sleek black cat with green eyes is resting on a quiet sandy beach at golden hour, static camera \\
Enumerated & Subject & 3 & A tall spotted male giraffe with a long neck is resting on a quiet sandy beach at golden hour, static camera \\
Enumerated & Subject & 4 & A chubby gray baby seal with whiskers is resting on a quiet sandy beach at golden hour, static camera \\
Enumerated & Subject & 5 & A colorful adult macaw with bright feathers and a curved beak is resting on a quiet sandy beach at golden hour, static camera \\
Enumerated & Scene & 1 & A large fluffy adult golden retriever is resting in a cozy sunlit living room with wooden floors, static camera \\
Enumerated & Scene & 2 & A large fluffy adult golden retriever is resting on a quiet snowy field under an overcast sky, static camera \\
Enumerated & Scene & 3 & A large fluffy adult golden retriever is resting in a narrow neon-lit alley at night, static camera \\
Enumerated & Scene & 4 & A large fluffy adult golden retriever is resting on a grassy hilltop during a breezy sunrise, static camera \\
Enumerated & Scene & 5 & A large fluffy adult golden retriever is resting inside a rustic barn with warm lantern light, static camera \\
Enumerated & Motion & 1 & A large fluffy adult golden retriever is slowly wagging its tail on a quiet sandy beach at golden hour, static camera \\
Enumerated & Motion & 2 & A large fluffy adult golden retriever is trotting along the shoreline on a quiet sandy beach at golden hour, static camera \\
Enumerated & Motion & 3 & A large fluffy adult golden retriever is digging energetically in the sand on a quiet sandy beach at golden hour, static camera \\
Enumerated & Motion & 4 & A large fluffy adult golden retriever is stretching its front legs forward on a quiet sandy beach at golden hour, static camera \\
Enumerated & Motion & 5 & A large fluffy adult golden retriever is shaking water off its fur on a quiet sandy beach at golden hour, static camera \\
Enumerated & Camera & 1 & A large fluffy adult golden retriever is resting on a quiet sandy beach at golden hour, slow push-in camera \\
Enumerated & Camera & 2 & A large fluffy adult golden retriever is resting on a quiet sandy beach at golden hour, gentle pull-back camera \\
Enumerated & Camera & 3 & A large fluffy adult golden retriever is resting on a quiet sandy beach at golden hour, smooth left-to-right tracking camera \\
Enumerated & Camera & 4 & A large fluffy adult golden retriever is resting on a quiet sandy beach at golden hour, slow orbiting camera \\
Enumerated & Camera & 5 & A large fluffy adult golden retriever is resting on a quiet sandy beach at golden hour, rising crane camera \\
\bottomrule
\end{tabular}
\caption{Complete controlled-prompt example for one semantic group. Only the target span changes within each intervention block.}
\label{tab:app5-prompt-case}

\end{table*}

\subsection{Statistical Details}
\label{app5:statistics}

For model $m$, semantic group $i$, target dimension $d$, and intervention $r$, we compute
\begin{equation}
\Delta_{m,i,d}^{(r)} = \mathrm{MPD}_{m,i,d}^{(r)}-\mathrm{MPD}_{m,i,d}^{(\mathrm{Specified})}.
\end{equation}
Model means average this paired difference over the 20 groups. For the panel-level analysis, we first average the seven fixed models within each group and then use the 20 group means as the statistical units. We report a 20,000-sample group bootstrap confidence interval, block win rate, paired Hedges $g_z$, Wilcoxon signed-rank $p$ values, and BH-adjusted $q$ values.
Full results are in Table \ref{tab:app5-group-effects} and \ref{tab:app5-contrasts}.

\begin{table*}[h]
\centering
\small
\begin{tabular}{@{}lllrrrr@{}}
\toprule
Intervention & Dimension & Mean $\Delta$ [95\% CI] & Positive blocks & $g_z$ & $p$ & $q$ \\
\midrule
Hinted & Subject & +0.197 [+0.144,+0.251] & 19/20 & 1.49 & $1.91\times10^{-6}$ & $3.00\times10^{-6}$ \\
Hinted & Scene & +0.056 [+0.032,+0.087] & 17/20 & 0.81 & $3.15\times10^{-5}$ & $4.33\times10^{-5}$ \\
Hinted & Motion & +0.069 [+0.044,+0.094] & 18/20 & 1.13 & $1.31\times10^{-4}$ & $1.60\times10^{-4}$ \\
Hinted & Camera & +0.210 [+0.184,+0.239] & 20/20 & 3.05 & $9.54\times10^{-7}$ & $1.75\times10^{-6}$ \\
Omitted & Scene & +0.051 [+0.025,+0.082] & 18/20 & 0.73 & $1.97\times10^{-4}$ & $2.17\times10^{-4}$ \\
Omitted & Motion & +0.011 [-0.007,+0.029] & 12/20 & 0.26 & 0.139 & 0.139 \\
Omitted & Camera & +0.092 [+0.069,+0.117] & 20/20 & 1.58 & $9.54\times10^{-7}$ & $1.75\times10^{-6}$ \\
Enumerated & Subject & +0.473 [+0.416,+0.528] & 20/20 & 3.42 & $9.54\times10^{-7}$ & $1.75\times10^{-6}$ \\
Enumerated & Scene & +0.370 [+0.304,+0.435] & 20/20 & 2.31 & $9.54\times10^{-7}$ & $1.75\times10^{-6}$ \\
Enumerated & Motion & +0.146 [+0.125,+0.170] & 20/20 & 2.66 & $9.54\times10^{-7}$ & $1.75\times10^{-6}$ \\
Enumerated & Camera & +0.321 [+0.288,+0.357] & 20/20 & 3.85 & $9.54\times10^{-7}$ & $1.75\times10^{-6}$ \\
\bottomrule
\end{tabular}
\caption{Complete group-level uncertainty for the target responses. This table supplements the model-level means in the main paper.}
\label{tab:app5-group-effects}

\end{table*}

\begin{table*}[h]
\centering
\scriptsize
\begin{tabular}{@{}llrrrr@{}}
\toprule
Contrast & Dimension & Mean difference [95\% CI] & Positive blocks & Positive models & $q$ \\
\midrule
Hinted $-$ Omitted & Scene & +0.005 [-0.007,+0.018] & 10/20 & 4/7 & 0.298 \\
Hinted $-$ Omitted & Motion & +0.058 [+0.032,+0.087] & 17/20 & 6/7 & $1.17\times10^{-4}$ \\
Hinted $-$ Omitted & Camera & +0.118 [+0.092,+0.147] & 20/20 & 6/7 & $1.00\times10^{-6}$ \\
Enumerated $-$ Hinted & Subject & +0.275 [+0.250,+0.301] & 20/20 & 7/7 & $1.00\times10^{-6}$ \\
Enumerated $-$ Hinted & Scene & +0.313 [+0.262,+0.364] & 20/20 & 7/7 & $1.00\times10^{-6}$ \\
Enumerated $-$ Hinted & Motion & +0.077 [+0.057,+0.098] & 20/20 & 7/7 & $1.00\times10^{-6}$ \\
Enumerated $-$ Hinted & Camera & +0.112 [+0.078,+0.143] & 19/20 & 5/7 & $1.20\times10^{-5}$ \\
\bottomrule
\end{tabular}
\caption{Adjacent-setting contrasts computed within each target dimension.}
\label{tab:app5-contrasts}

\end{table*}

\subsection{Faithfulness Evaluation and Realization Cost}
\label{app5:faithfulness}

We compare Enumerated with Hinted within the same model, semantic group, and target dimension. Since Hinted prompts are less specific and therefore easier to satisfy, this contrast measures the realization cost of adding explicit control. Results are reported in Table \ref{tab:app5-faithfulness}.

\begin{table*}[h]
\centering
\small
\resizebox{\textwidth}{!}{%
\begin{tabular}{@{}lrrrrrr@{}}
\toprule
Dimension & Hinted mean & Enumerated mean & Enumerated $-$ Hinted [95\% CI] & Negative blocks & Negative models & $q$ \\
\midrule
Subject & 4.593 & 4.497 & -0.096 [-0.258,+0.048] & 10/20 & 5/7 & 0.226 \\
Scene & 4.999 & 4.791 & -0.204 [-0.254,-0.156] & 19/20 & 7/7 & $8.90\times10^{-5}$ \\
Motion & 4.780 & 4.090 & -0.690 [-0.887,-0.486] & 18/20 & 7/7 & $3.82\times10^{-5}$ \\
Camera & 4.469 & 4.075 & -0.393 [-0.541,-0.247] & 18/20 & 6/7 & $8.90\times10^{-5}$ \\
\bottomrule
\end{tabular}
}%
\caption{Faithfulness changes when moving from Hinted to explicit enumeration.}
\label{tab:app5-faithfulness}

\end{table*}

\subsection{Failure Modes Revealed by Controlled Prompting}
\label{app5:failure-modes}

\begin{figure*}[h]
\centering
\includegraphics[width=\linewidth]{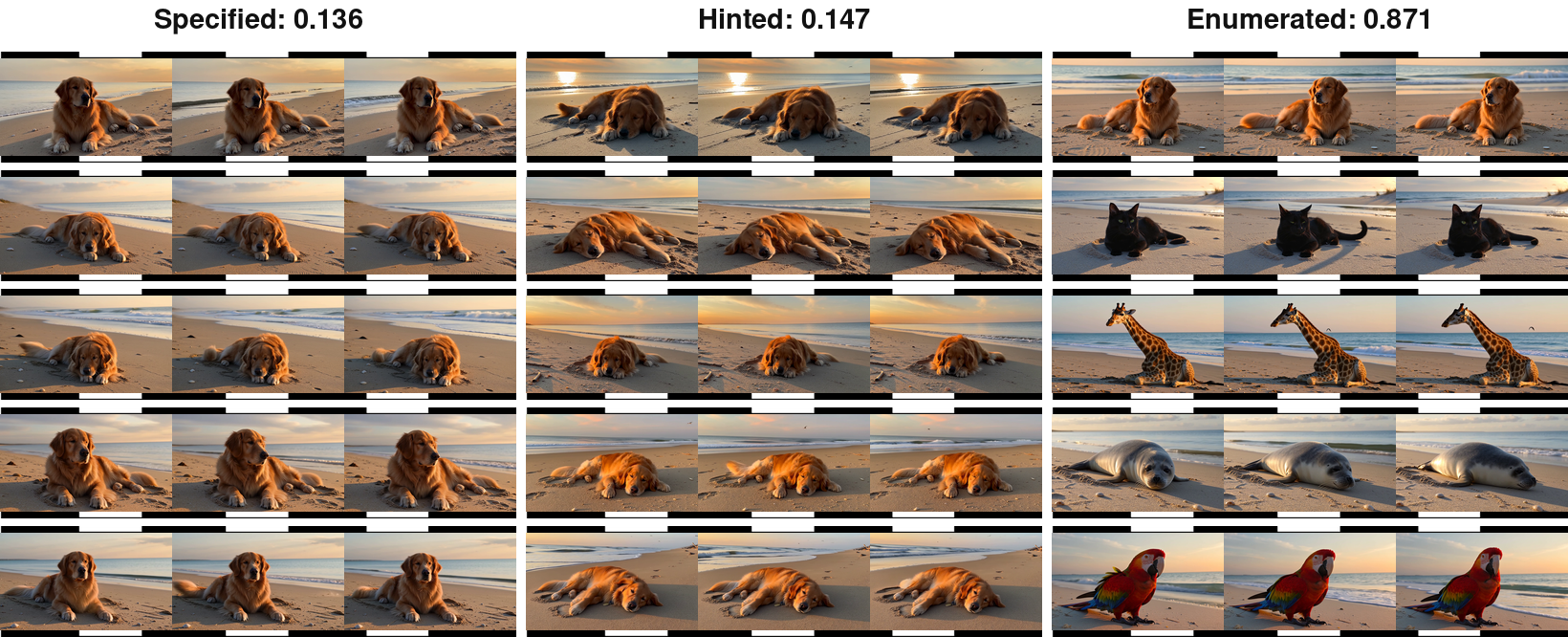}
\caption{Prompt-gated default convergence.}
\label{fig:app5-default-convergence}
\end{figure*}

\paragraph{Prompt-gated default convergence.}
When a target dimension is replaced by a broad cue, the model may preserve a narrow default realization. In the Subject case (Figure \ref{fig:app5-default-convergence}), HappyHorse produces mostly the same dog under \texttt{an animal}: Specified, Hinted, and Enumerated MPD are 0.136, 0.147, and 0.871.

\begin{figure*}[h]
\centering
\begin{minipage}[t]{0.49\textwidth}
\centering
\includegraphics[width=\linewidth]{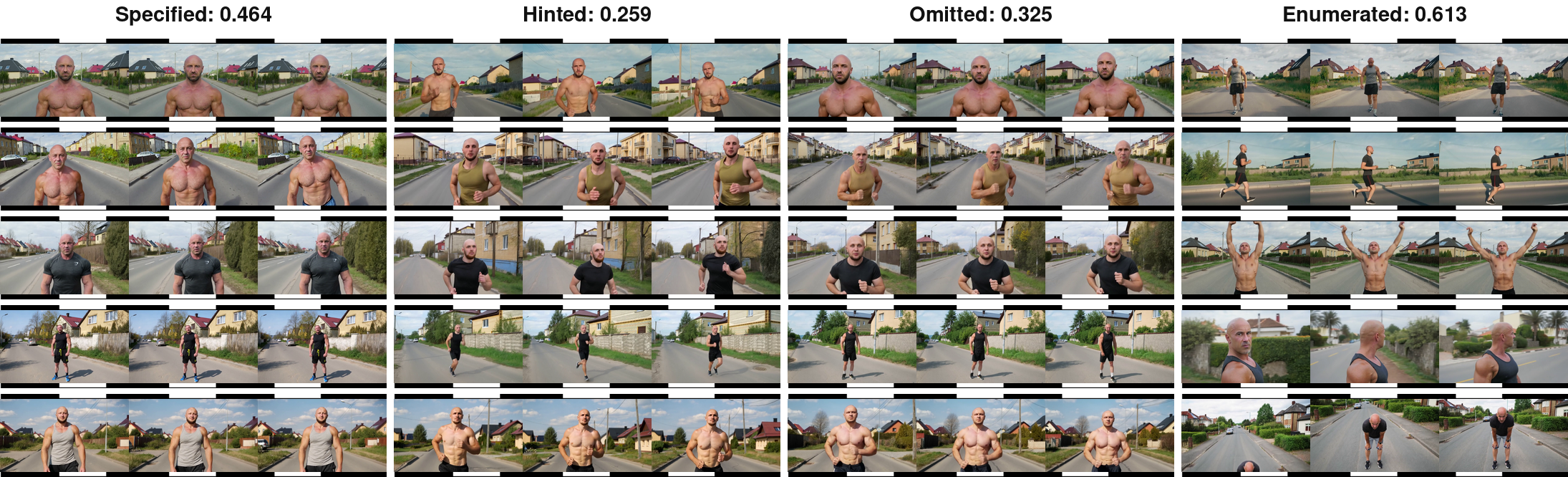}
\end{minipage}\hfill
\begin{minipage}[t]{0.49\textwidth}
\centering
\includegraphics[width=\linewidth]{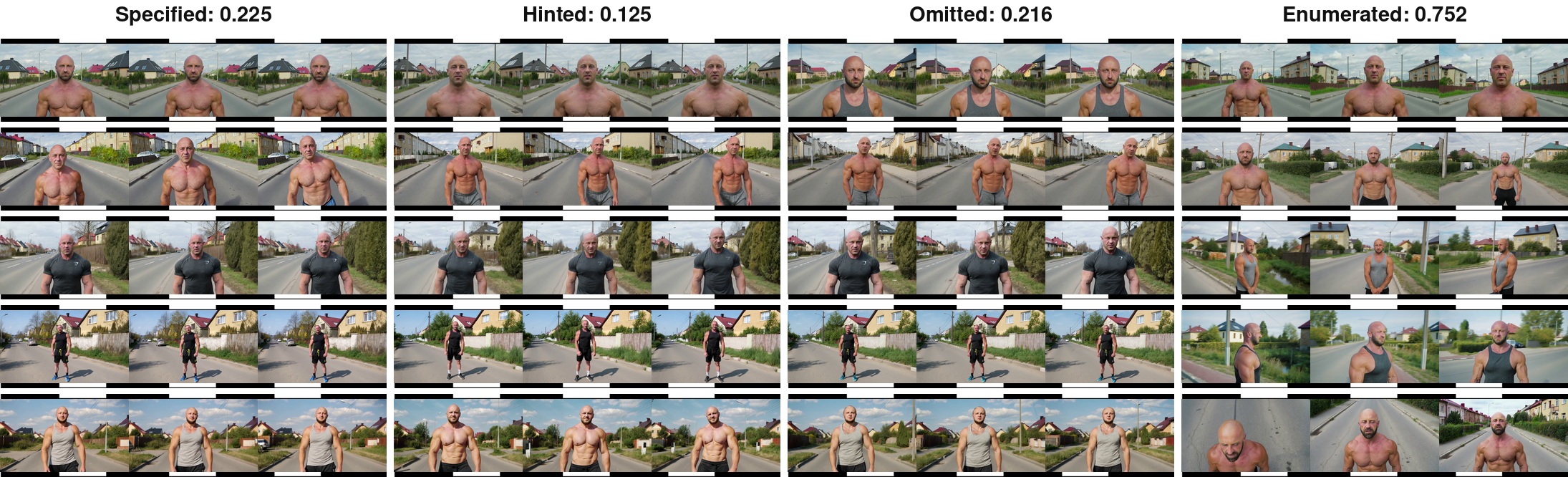}
\end{minipage}
\caption{Both cases use Hunyuan; the left column releases Motion and the right column releases Camera.}
\label{fig:app5-dynamic-underexploration}
\end{figure*}

\paragraph{Dynamics.}
The same diagnostic appears in temporal dimensions (Figure \ref{fig:app5-dynamic-underexploration}). For Hunyuan on the road-person prompt, Motion MPD falls from 0.464 in Specified to 0.259 with a generic cue and 0.325 after omission, but rises to 0.613 under explicit action candidates. Camera shows the same pattern in the matched block: 0.225, 0.125, 0.216, and 0.752 for Specified, Hinted, Omitted, and Enumerated.

\begin{figure*}[h]
\centering
\begin{minipage}[t]{0.49\textwidth}
\centering
\includegraphics[width=\linewidth]{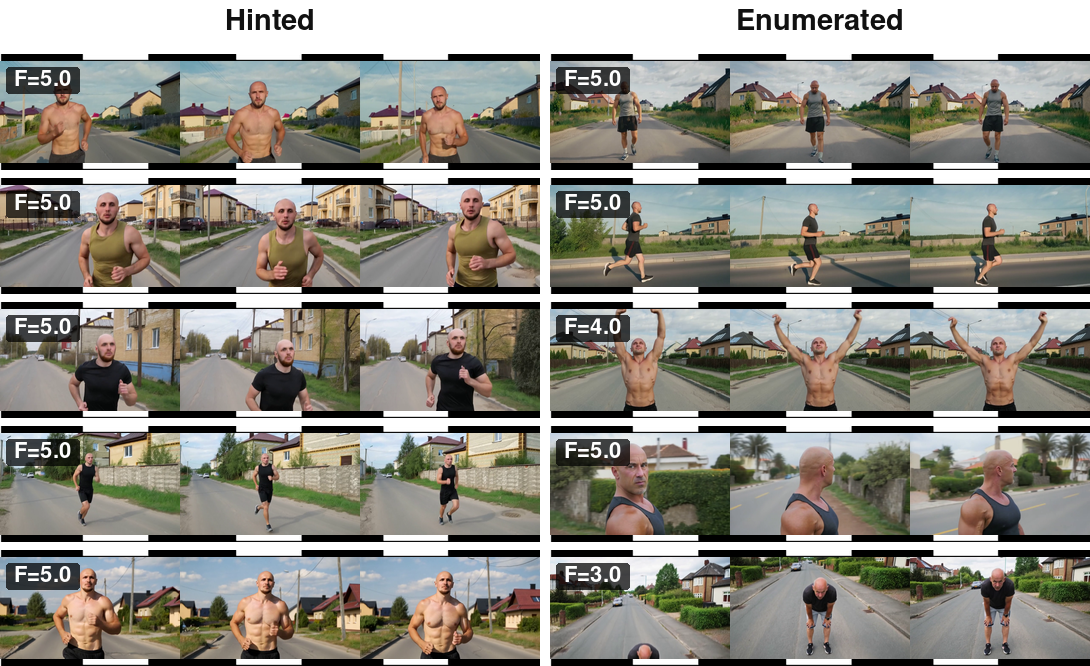}
\end{minipage}\hfill
\begin{minipage}[t]{0.49\textwidth}
\centering
\includegraphics[width=\linewidth]{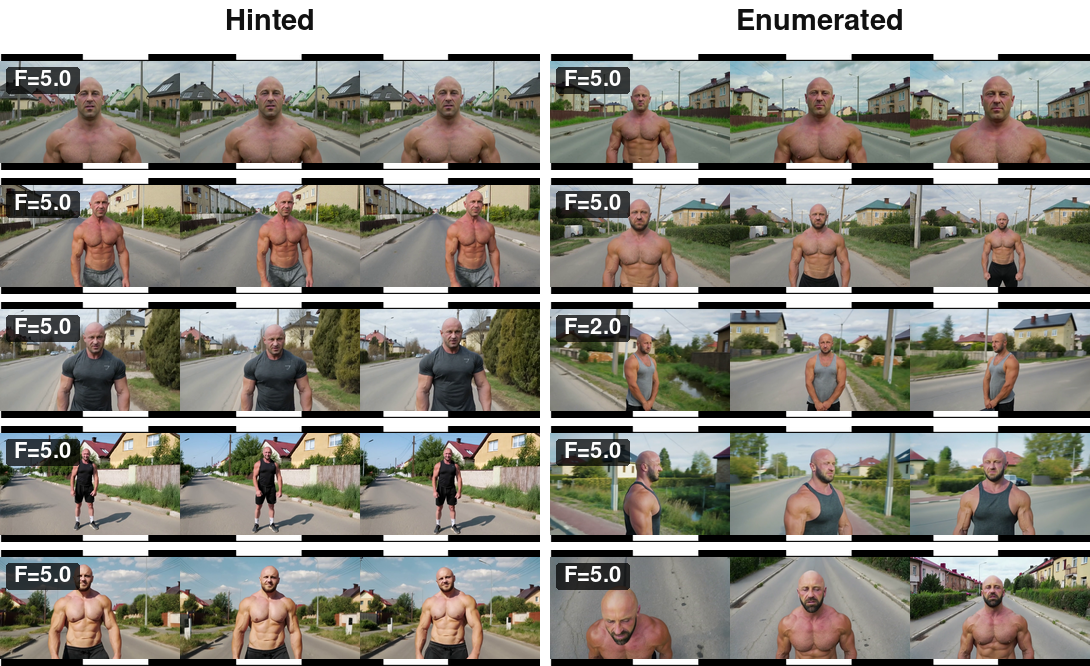}
\end{minipage}
\caption{Realization-limited diversity. Left: Hunyuan Motion. Right: Hunyuan Camera. Each row displays the target-dimension Faithfulness score on the video.}
\label{fig:app5-realization-cases}
\end{figure*}

\paragraph{Realization-limited diversity.}
Explicit enumeration can enlarge the measured span while some candidates remain only partially realized. In the Hunyuan Motion case (Figure \ref{fig:app5-realization-cases}), Hinted videos all receive $F=5$, whereas the five Enumerated candidates include scores 5, 5, 4, 5, and 3. In the corresponding Camera case, the Enumerated scores include one $F=2$ despite a large MPD gain. These cases motivate reporting diversity and faithfulness separately.

\begin{figure*}[h]
    \centering
    \includegraphics[width=1.0\linewidth]{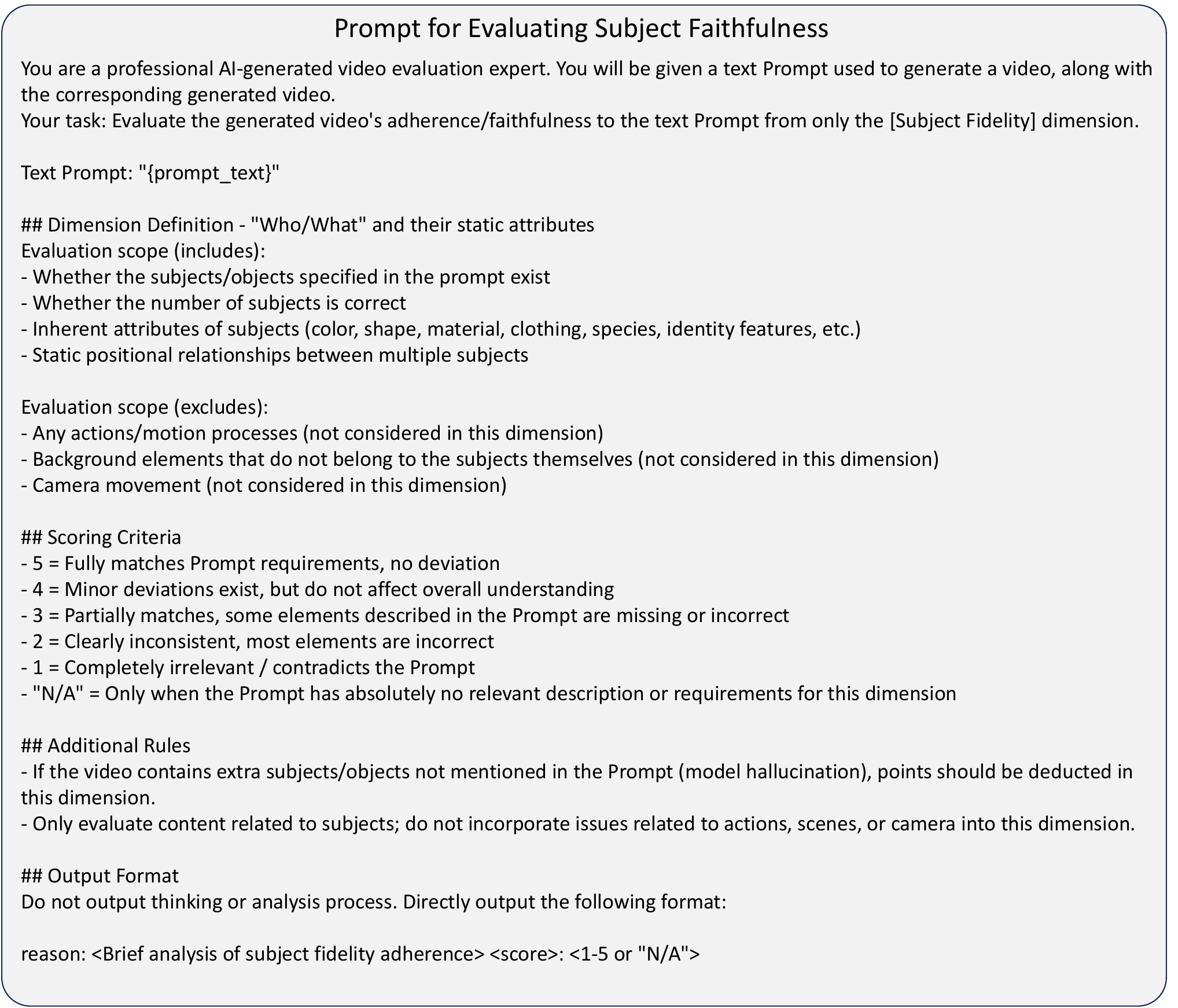}
    \caption{Prompt for Evaluating Subject Faithfulness.}
    \label{fig:faith_subject}
\end{figure*}

\begin{figure*}[h]
    \centering
    \includegraphics[width=1.0\linewidth]{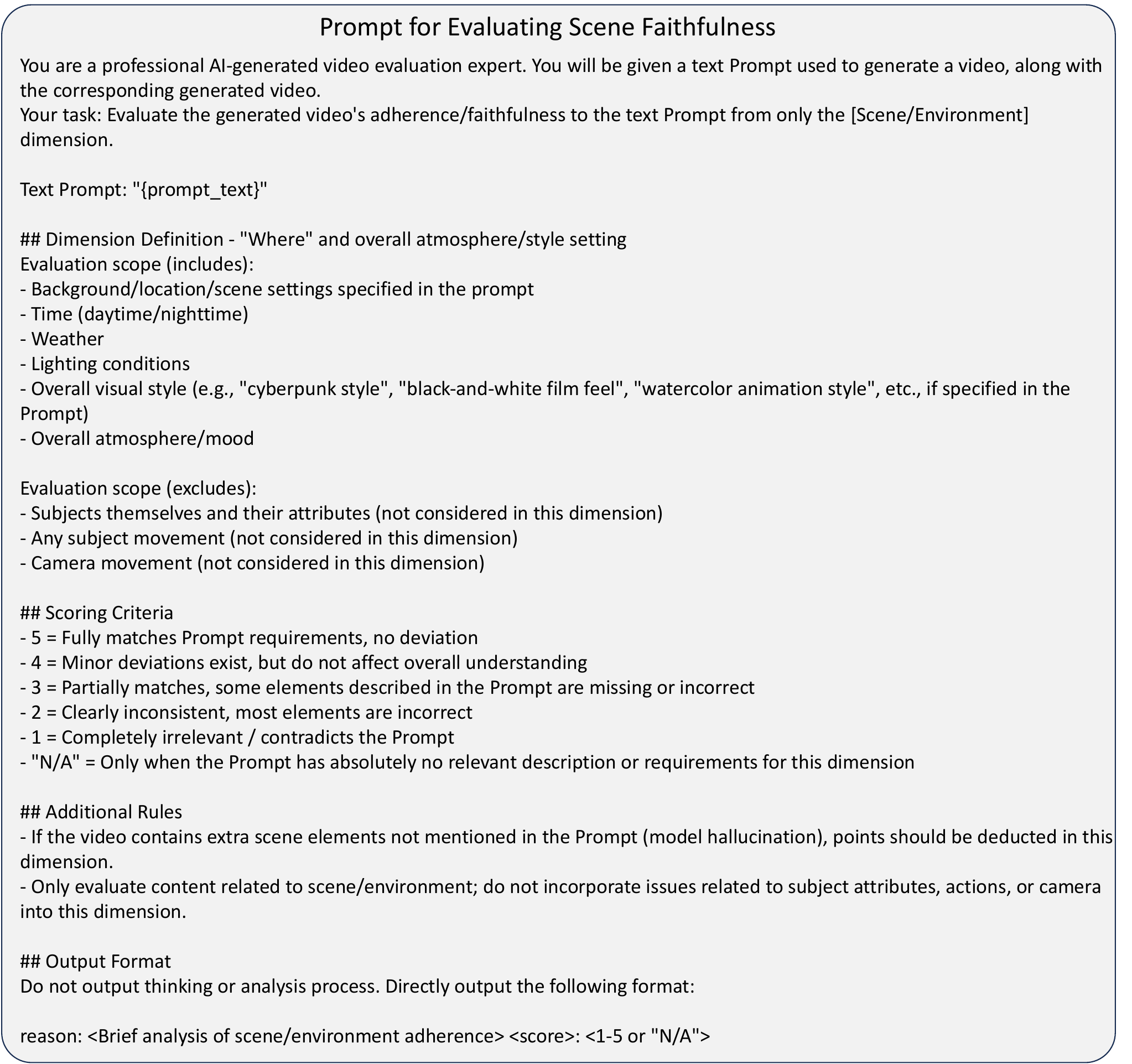}
    \caption{Prompt for Evaluating Scene Faithfulness.}
    \label{fig:faith_scene}
\end{figure*}

\begin{figure*}[h]
    \centering
    \includegraphics[width=1.0\linewidth]{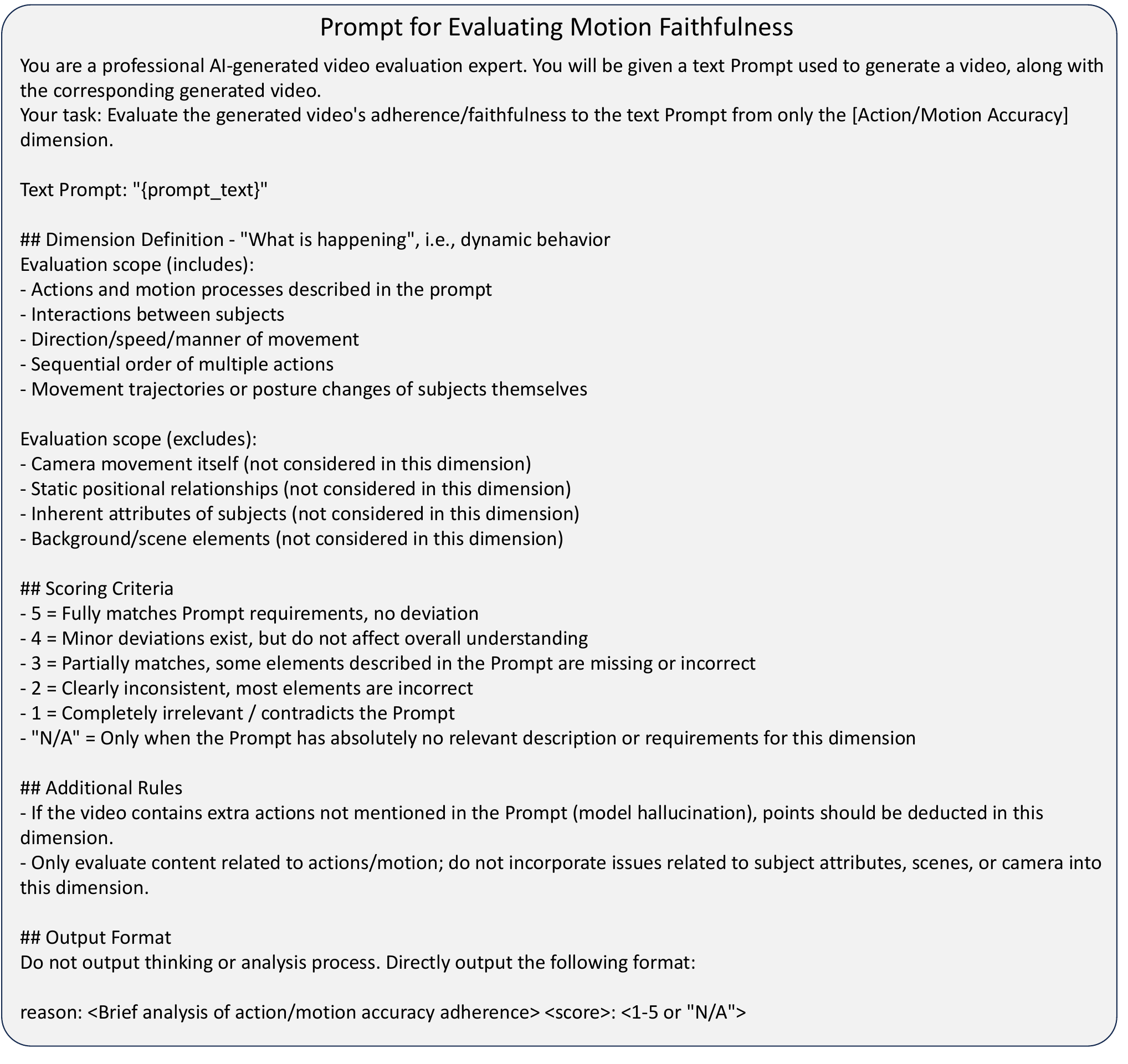}
    \caption{Prompt for Evaluating Motion Faithfulness.}
    \label{fig:faith_motion}
\end{figure*}

\begin{figure*}[h]
    \centering
    \includegraphics[width=1.0\linewidth]{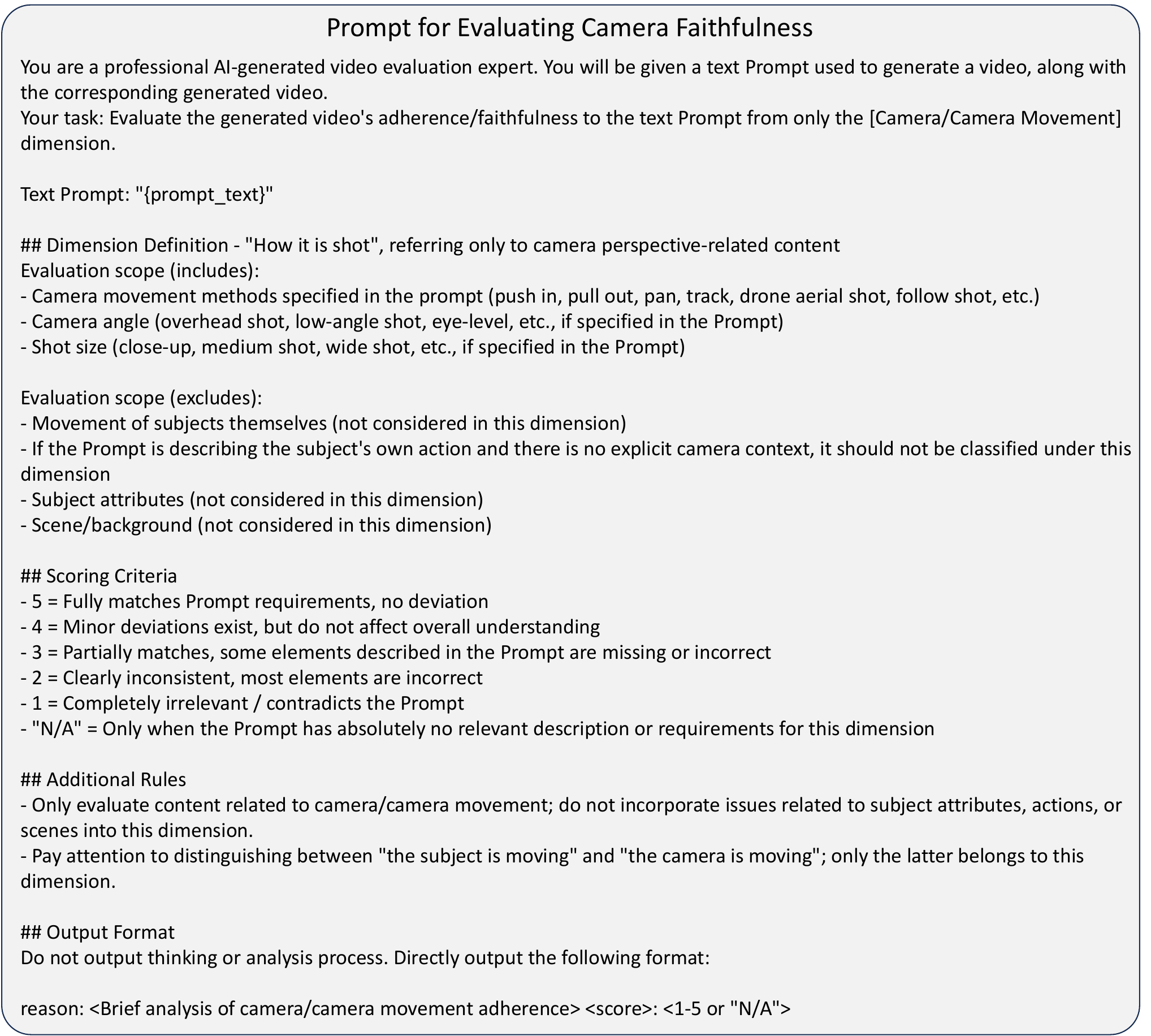}
    \caption{Prompt for Evaluating Camera Faithfulness.}
    \label{fig:faith_camera}
\end{figure*}

\end{document}